\pdfoutput=1
\documentclass[conference]{IEEEtran}
\IEEEoverridecommandlockouts
\usepackage{cite}
\usepackage{amsmath,amssymb,amsfonts}
\usepackage{algorithmic}
\usepackage{graphicx} 
\usepackage{bmpsize}
\usepackage{textcomp}
\usepackage{xcolor}
\usepackage{booktabs} 
\def\BibTeX{{\rm B\kern-.05em{\sc i\kern-.025em b}\kern-.08em
    T\kern-.1667em\lower.7ex\hbox{E}\kern-.125emX}}

\usepackage{subfigure}
\usepackage{balance}

\usepackage{enumitem,kantlipsum}
\usepackage{tabularx}
\usepackage[flushleft]{threeparttable}

\usepackage{listings}
\usepackage{hyperref}
\hypersetup{
    colorlinks=true,
    linkcolor=blue,
    filecolor=magenta,
    urlcolor=black,
    }

\newcommand{\changes}[1]{{#1}}
\usepackage{ulem}
\DeclareRobustCommand{\erase}{\bgroup\markoverwith{\textcolor{red}{\rule[.5ex]{2pt}{0.4pt}}}\ULon}

\newcommand{\mlhpc}[1]{{#1}}

\newcommand{\dropcontent}[1]{}

\newcommand{\Lukas}[1]{\textcolor{cyan}{}}

\usepackage[symbol]{footmisc}

\usepackage{tablefootnote}
\usepackage{enumitem}

\usepackage{color}
\usepackage{listings}
\begin{document}

\title{
MLPerf\textsuperscript{TM} HPC: A \mlhpc{Holistic Benchmark Suite for Scientific} Machine Learning on HPC Systems
}

%%%%%%%%%%%%%
%% adding authors list
%%%%%%%%%%%%

\author{\IEEEauthorblockN{ 
  Steven Farrell \IEEEauthorrefmark{1},
  Murali Emani \IEEEauthorrefmark{2},
  Jacob Balma \IEEEauthorrefmark{3},
  Lukas Drescher \IEEEauthorrefmark{4},
  Aleksandr Drozd \IEEEauthorrefmark{5}, 
  Andreas Fink \IEEEauthorrefmark{4},\\ 
  Geoffrey Fox \IEEEauthorrefmark{6}, 
  David Kanter \IEEEauthorrefmark{7},
  Thorsten Kurth \IEEEauthorrefmark{8},
  Peter Mattson \IEEEauthorrefmark{9},
  Dawei Mu \IEEEauthorrefmark{10}, 
  Amit Ruhela \IEEEauthorrefmark{11}, \\
  Kento Sato \IEEEauthorrefmark{5},
  Koichi Shirahata \IEEEauthorrefmark{12},
  Tsuguchika Tabaru \IEEEauthorrefmark{12},
  Aristeidis Tsaris \IEEEauthorrefmark{13}, 
  Jan Balewski \IEEEauthorrefmark{1},
  Ben Cumming \IEEEauthorrefmark{4},\\
  Takumi Danjo \IEEEauthorrefmark{12},
  Jens Domke \IEEEauthorrefmark{5},
  Takaaki Fukai \IEEEauthorrefmark{5},
  Naoto Fukumoto \IEEEauthorrefmark{12},
  Tatsuya Fukushi \IEEEauthorrefmark{12},
  Balazs Gerofi \IEEEauthorrefmark{5},\\
  Takumi Honda \IEEEauthorrefmark{12},
  Toshiyuki Imamura  \IEEEauthorrefmark{5},
  Akihiko Kasagi \IEEEauthorrefmark{12},
  Kentaro Kawakami \IEEEauthorrefmark{12},
  Shuhei Kudo \IEEEauthorrefmark{5},\\
  Akiyoshi Kuroda \IEEEauthorrefmark{5},
  Maxime Martinasso  \IEEEauthorrefmark{4},
  Satoshi Matsuoka \IEEEauthorrefmark{5},
  Henrique Mendonça \IEEEauthorrefmark{4},
  Kazuki Minami \IEEEauthorrefmark{5},\\
  Prabhat Ram \IEEEauthorrefmark{14},
  Takashi Sawada \IEEEauthorrefmark{12},
  Mallikarjun Shankar \IEEEauthorrefmark{13},
  Tom St. John \IEEEauthorrefmark{15},
  Akihiro Tabuchi \IEEEauthorrefmark{12},\\
  Venkatram Vishwanath \IEEEauthorrefmark{2},
  Mohamed Wahib \IEEEauthorrefmark{16},
  Masafumi Yamazaki \IEEEauthorrefmark{12},
  Junqi Yin \IEEEauthorrefmark{13}
  }\\
  
\IEEEauthorrefmark{1}Lawrence Berkeley National Laboratory, USA
% Berkeley, CA 94720, USA \\
\IEEEauthorrefmark{2}Argonne National Laboratory, USA
% Lemont, IL 60439, USA \\
\IEEEauthorrefmark{3} Maxwell Labs, USA \\
\IEEEauthorrefmark{4} Swiss National Supercomputing Centre, Switzerland
% Via Trevano 131, 6900 Lugano,   
\IEEEauthorrefmark{5} RIKEN Center for Computational Science, Japan \\
% Kobe, Hyogo 650-0047, Japan \\ 
\IEEEauthorrefmark{6} University of Virginia, USA % Charlottesville, VA 22911, USA 
\IEEEauthorrefmark{7} MLCommons, USA 
\IEEEauthorrefmark{8} NVIDIA, Switzerland
\IEEEauthorrefmark{9} Google, USA \\
\IEEEauthorrefmark{10} National Center for Supercomputing Applications, USA
% Urbana, IL 61801, USA 
\IEEEauthorrefmark{11}  Texas Advanced Computing Center, USA \\
%Austin, TX 78758, USA \\
\IEEEauthorrefmark{12} Fujitsu Limited, Japan 
\IEEEauthorrefmark{13} Oak Ridge National Laboratory, USA 
% Oak Ridge, TN 37830, USA
\IEEEauthorrefmark{14} Microsoft, USA \\
\IEEEauthorrefmark{15} Cruise, USA 
\IEEEauthorrefmark{16} National Institute of Advanced Industrial Science and Technology, Japan
}

\maketitle

\begin{abstract}
\Lukas{Title discussion} Scientific communities are increasingly adopting machine learning and deep learning 
models in their applications to accelerate scientific insights. High performance computing systems are pushing the frontiers of performance with a rich diversity of hardware resources and massive scale-out capabilities. There is a critical need to understand fair and effective benchmarking of machine learning applications that are representative of real-world scientific use cases.
MLPerf\textsuperscript{\tiny{TM}} is a community-driven standard to benchmark machine learning workloads, focusing on end-to-end performance metrics. In this paper, we introduce MLPerf HPC, a benchmark suite of large-scale scientific machine learning training applications, driven by the MLCommons\textsuperscript{\tiny{TM}} Association. 
\mlhpc{We present the results from the first submission round including a diverse set of some of the world's largest HPC systems. We develop a systematic framework for their joint analysis and compare them 
in terms of data staging, algorithmic convergence and compute performance. As a result, we gain a quantitative understanding of optimizations on different subsystems such as staging and on-node loading of data, compute-unit utilization and communication scheduling enabling overall 
$>10 \times$ (end-to-end) 
performance improvements through system scaling.
Notably, our analysis shows a scale-dependent interplay between the dataset size, a system's memory hierarchy and training convergence 
that underlines the importance of near-compute storage.
To overcome the data-parallel scalability challenge at large batch-sizes, 
we discuss specific learning techniques and hybrid data-and-model parallelism that are effective on large systems. %on some of the largest systems in this round. 
We conclude %this paper 
by characterizing each benchmark with respect to low-level memory, I/O and network behaviour to parameterize extended roofline performance models in future rounds.}
\end{abstract}

\setlength{\abovedisplayskip}{3pt}
\setlength{\belowdisplayskip}{3pt}

\vspace{-0.05in}
\section{Introduction}
Scientific applications are leveraging the potential of machine learning (ML) \mlhpc{and, especially, deep learning} to accelerate scientific discovery. This trend is prevalent in multiple domains, such as cosmology, particle physics, biology and clean energy.
These applications are innately distinct from traditional industrial applications with respect to the type and volume of data and the resulting model complexity.
Recently, ML-driven applications have led to major insights and scientific discoveries that would have taken years using traditional methods. A recent breakthrough in addressing one such grand challenge in biology was the development of an ML model that solved the protein folding problem with the AlphaFold tool by Alphabet's Deepmind~\cite{deepmind}. 
Large-scale experiments will yield unprecedented volumes of scientific data. The authors of~\cite{ai4sciencechallenges} highlight the challenges faced with understanding the exponential growth of experimental data and present opportunities of using machine learning techniques to advance science.
The AI for Science report~\cite{AIforScience} put forth by stakeholders from leadership computing facilities, DOE labs, and academia details a vision for leveraging AI in science applications critical to US DOE missions. The vision outlined in this report emphasizes the need for a systematic understanding of how these applications perform on diverse supercomputers.

An ideal benchmark suite will help assess HPC system performance while 
driving innovation in systems and methods.
Developing one is difficult because of the inherent trade-off between building generalizable and representative proxies of real ML workloads, and building high-fidelity proxies that probe specific features of an HPC system. Benchmarks of the former type are general enough to capture what the average user on a large multi-user system is doing with AI  while benchmarks of the latter type focus on how a specific kernel performs on the system.
Implementation of ML models on supercomputers at full scale poses challenges that are not typically exposed at small-scale, such as the I/O impact of large-scale datasets.   
Hence, adopting existing benchmarking approaches for scientific machine learning problems would not be able to capture realistic behaviour of the scientific applications. 

There have been significant efforts in benchmarking supercomputers with traditional HPC workloads with major ones listed in Table \ref{tab:existing-work}.
TOP500~\cite{Top500Nov2020} ranks supercomputers across the world and publishes their performance numbers (in Flops) 
with High Performance Linpack (HPL).
It captures many of the general features that large-scale scientific applications share, such as domain decomposition and heavy use of linear algebra solvers. 
A single benchmark can be run across the entire system in a weak-scaling fashion.
Green500~\cite{green500} ranks supercomputers based on their energy efficiency. The list reports the performance per rated watt using the LINPACK benchmark at double precision format. 
\begin{table*}[htbp]
\footnotesize
    \centering
    \caption{HPC Machine Learning Benchmarks}
    \begin{tabular}{l p{1.25cm} p{1.75cm} p{1.2cm} p{10cm}}
        \toprule
        \textbf{Benchmark} & \textbf{Performance metrics} & \textbf{Application domain} & \textbf{Data volume} & \textbf{Comments}\\
        \midrule
        HPL, HPL-AI & Flops, Flops/Watt &  Random dense system of linear equations & Variable & Used in Top500 and Green500 to rank supercomputers. Problem size scaled to optimize the performance for machine size. HPL measures performance at double precision, HPL-AI measures  performance in mixed precision\\
        \hline
        HPCAI500 & Valid Flops, Valid Flops/Watt & Image classification, Weather analytics & 150 GB \& 1.65 TB & Convolution and GEMM layers measure the performance in valid Flops which impose penalty based on failure to meet target accuracy. Limited to Microbenchmarks, Object Detection and Image Classification tasks with microscopic view of common deep learning models (Faster-RCNN, ResNet) \\
        \hline
        Deep500 & Throughput, Time to solution & any machine learning application & 150 GB %(ImageNet) 
        &  Provides infrastructure to help evaluate different framework implementations and multiple levels of operators. Challenging to integrate into scientific applications. Evaluated with ImageNet dataset.\\
        \hline
        MLPerf HPC & Time to train & Cosmology and weather analytics & 5.1 TB \& 8.8 TB & Targets representative scientific machine learning applications with massive datasets. Provision of two types submissions, closed and open enable novel optimizations. Time to solution metric and focused timing captures holistic model performance\\
        \bottomrule
    \end{tabular}
    \label{tab:existing-work}
    \vspace{-.225in}
\end{table*}
Several benchmarking efforts have previously aimed to characterize performance of machine learning workloads, including Deep500~\cite{Deep500},
HPCAI500~\cite{HPCAI500}, 
and HPL-AI~\cite{HPLAI}. The largest dataset used across these attempts is obtained from the Extreme Weather Dataset~\cite{extremeweatherdataset} of about 1.65 TB. 
Other benchmarks aimed at analyzing model performance include DAWNBench~\cite{dawnbench}, 
DeepBench~\cite{deepbench}, Fathom~\cite{Fathom}, ParaDNN~\cite{paradnn}, HPE DLBS~\cite{hpe}, and XSP~\cite{XSP}.
The challenges and limitations of existing benchmarking efforts drive the need to develop a benchmark suite with science applications that run at scale.

In this paper, we present MLPerf HPC, a benchmark suite
aimed to address the limitations of prior efforts. \changes {This benchmark suite is a part of MLPerf,} driven by MLCommons~\cite{mlcommons}, an open engineering consortium that aims to accelerate machine learning innovation through MLPerf benchmarks, public datasets, and best practices. 
MLPerf Training benchmarks~\cite{mlperf-training} aim to measure the performance of training models while MLPerf Inference benchmarks~\cite{mlperf-inference} aim to measure how fast systems can produce results using a trained model. MLCommons takes a neutral stand about any form of comparison of results across submissions. 

\noindent The primary contributions of this work are:
\begin{enumerate}[nosep,leftmargin=*]
\item 
\changes{Introduce the  MLPerf HPC benchmark suite, the newest ML training benchmark suite from the MLCommons consortium, and the first to specifically focus on scientific ML applications relevant for HPC systems (section \ref{sec:benchmark-suite}). We will describe the first two benchmark applications, CosmoFlow and DeepCAM, as well as the benchmarking methodology and process (section \ref{sec:benchmarking-process}).}
% \item Discuss the two scientific machine learning applications of the MLPerf HPC benchmark suite, CosmoFlow and DeepCAM, and describe the benchmarking methodology and process.
\item \changes{Present results from the inaugural  MLPerf HPC submission round in 2020 (section \ref{sec:results}). These results feature measurements from leading supercomputing platforms around the world, innovations in scalable model-and-data-parallel training and learning algorithms, and the largest scale  MLPerf submission to date.} 
\item \changes{Develop a systematic framework for the joint analysis of the publicly available  MLPerf HPC submission results to clearly understand and compare submissions in terms of data staging, algorithmic convergence and system compute performance in a condensed set of plots that enable new insights (subsection \ref{ssec:results-analysis}).}
\item \changes{Gain a quantitative understanding of how system-level optimizations lead to $>10\times$ improvements in end-to-end performance for both benchmarks (closed division), what constrains these submissions from scaling further and discuss alternative learning techniques that enable further improvements by a factor of 1.1-3.4$\times$ (open division, subsection \ref{ssec:highlights}).}
\item 
\changes{Introduce techniques to characterize memory, network, and I/O behaviours of the benchmark applications across all involved systems, relate them to  MLPerf HPC submission results and, hereby, lay the groundwork for a future characterization of the submissions in extended roofline performance models \cite{cardwell2019extended} (section \ref{sec:workload-characterization}).}

\end{enumerate}

\vspace{-0.05in}
\section{MLPerf HPC Benchmark Suite}
\label{sec:benchmark-suite}
The MLPerf HPC benchmarks are
holistic in the sense that they capture critical features of emerging scientific machine learning workloads: massive data volume, large complex models, and 
training schemes that incorporate data-parallelism, model-parallelism, and hyper-parameter optimization.
The goal is to drive innovation in HPC system and software design for machine learning workloads, especially those applications 
that depend heavily on accelerator devices for fast compute, interconnects for high-bandwidth, low-latency communication, or I/O subsystems that govern the rate at which data can be accessed. 
Additional requirements of the benchmark suite, such as training to convergence, profile generation, and coarse-grained time reporting enable each individual benchmark's performance to be characterized relative to its utilization of a system's I/O, communication, memory, and compute capabilities. This makes the MLPerf HPC benchmark suite uniquely capable of characterizing the ability of existing HPC systems to run an exciting class of new workloads, while simultaneously providing engineers a 
standard set of benchmarks for informing the design of tomorrow's large-scale high performance computers.
\changes{The MLPerf HPC benchmarks are also unique in that they offer performance characterization capability for state-of-the-art practices in scientific discovery. 
The emerging domain of coupling simulations with AI 
motivates the first two benchmarks included in the suite, CosmoFlow and DeepCAM. The reference implementations of these applications are available at \cite{mlperfhpc}. Both benchmarks incorporate massive datasets based on simulations and predict important parameters with unprecedented accuracy in their respective domains - cosmology and extreme weather. Details for each benchmark is provided in the following subsections.

Future benchmarks for the suite will be community driven, aiming to share similar characteristics with CosmoFlow and DeepCAM while also expanding the relevance of the benchmark suite to other scientific regimes where AI is quickly becoming an important tool for discovery, such as computational biology, materials science and personalized medicine. }

\vspace{-1mm}
\subsection{\textbf{CosmoFlow}}
\label{ssec:benchmark-suite-cosmoflow}
The CosmoFlow benchmark is based on the work by Mathuriya et. al.~\cite{cosmoflow}, continued by the ECP ExaLearn project~\cite{exalearn}. The task is to predict cosmological parameters from the distribution and structure of dark matter in the universe. The dataset comes from N-body cosmology simulations produced by the ExaLearn team~\cite{cosmo-dataset} binned into 3D volumetric histograms of size $512^3$ with four channels representing different red-shift values. These massive volumes present considerable challenges for training models due to large memory footprint, and so, similar to what is done in~\cite{cosmoflow}, the samples are split into smaller cubes of size $128^3$ with four red-shift channels. The target quantities are four cosmology parameters, $\Omega_M$, $\sigma_8$, $n_s$, and $H_0$, which are important to describe the evolution of the universe. The final dataset used for this benchmark has 262,144 samples for training and 65,536 samples for testing and is stored in TFRecord~\cite{tfrecord} files.

\begin{table}
\footnotesize
    \centering
    \caption{MLPerf HPC Benchmarks Overview}
    \begin{tabular}{cccp{3cm}}
        \toprule
        \textbf{Benchmark} & \textbf{Quality target} & \textbf{\#Runs} &
                \begin{tabular}{c}\textbf{Tunable} \\ \textbf{hyperparameters} \end{tabular} \\
            
           % \textbf{Modifiable Hyperparameters} \\
        \midrule
        CosmoFlow & MAE $<$ 0.124 & 10 & batch size, learning rates \\
        DeepCAM & IOU $>$ 0.82 & 5 & optimizer (LAMB or AdamW), \\
        & & & batch size, learning rates \\
        \bottomrule
    \end{tabular}
    \label{tab:benchmarks-overview}
    \vspace{-6mm}
\end{table}

The CosmoFlow reference model was adapted from~\cite{jan-cosmoflow} which introduced some modifications with respect to the original published work. The model is a 3D convolutional neural network with five convolutional layers and three fully-connected layers. Each convolutional layer has kernel size 2 with $32 \times i$ filters in the $i$th layer. The first two fully connected layers have sizes $128$ and $64$, respectively. The final layer has output size $4$, corresponding to the predicted target quantities. All hidden layers have leaky ReLU activations, with the exception of the output layer which has a tanh activation scaled by a factor $1.2$. After each convolutional layer, there is a 3D Max-Pool operation reducing the sample size by half along each dimension. Finally, the model has dropout layers after the first two fully connected layers with dropout probability $0.5$. The model is trained with a mean-squared-error (MSE) loss function and the standard SGD optimizer with \mlhpc{a baseline learning rate schedule consisting of} an initial learning rate of $0.001$ which is dropped to $2.5 \times 10^{-4}$ at 32 epochs and $1.25 \times 10^{-4}$ at 64 epochs. The \mlhpc{baseline} global batch size is set to 64. \mlhpc{To scale above this configuration, the reference implementation multiplies the baseline learning rate by the factor of batch size increase, respectively its square-root.} 

The target quality is chosen to be mean-absolute-error (MAE) $<$ 0.124 %, which allows for good convergence 
when scaling the batch size and learning rate above the reference configuration. CosmoFlow training %was shown to 
exhibited high variability in the number of epochs to converge, which motivated a requirement of 10 training runs to get a reliable measurement of the time to train.
\vspace{-2mm}
\subsection{\textbf{DeepCAM}}
\label{ssec:benchmark-suite-deepcam}
DeepCAM \cite{Deepcam} implements a convolutional encoder-decoder segmentation architecture trained on CAM5 climate simulation data \cite{cam5} with TECA generated heuristic segmentation masks~\cite{PRABHAT2012866} to identify extreme weather phenomena such as atmospheric rivers and tropical cyclones.
DeepCAM was the first deep learning application which scaled to the full OLCF Summit system \cite{Summit} and was awarded the ACM Gordon Bell Prize in 2018~\cite{kurth2018exascale}. 
Since then, the model was developed into its current form: the ResNet-50~\cite{he2015deep} encoder backend was replaced with an Xception~\cite{chollet2017xception} network,
batch normalization was re-introduced and LAMB~\cite{you2020large} replaced the original ADAM/LARS optimizer~\cite{kingma2017adam,you2017large}.
The most notable features of this network are 20 residual blocks comprised of depthwise-separable convolutions, which are themselves comprised of grouped convolutions with maximal group count, followed by a point-wise convolution. The bottleneck layer employs atrous spatial pyramid pooling~\cite{chen2017deeplab} with various filter sizes, and global average pooling to extract features at different scales. The results of those operations are concatenated and fed to the deconvolutional decoder. Outside the residual blocks, the network has a single skip connection which propagates low level features directly to the decoder without routing them through the bottleneck layer.

The network takes $16 \times 1152 \times 768$ sized input tensors and predicts $1152 \times 768$ sized segmentation masks for three classes (background, tropical cyclone/hurricane, atmospheric river). There are 121,266 training and 15,158 testing samples and no data augmentation is used. 
DeepCAM is trained with weighted cross-entropy loss, to account for the high class imbalance (about 95\% of the pixels are background). The target score is the intersection-over-union (IOU)
between the predictions and the targets. The scientifically motivated target score is 0.82, which corresponds to a similarity of 82\%.

\vspace{5mm} %5mm vertical space
It is critical to understand what differentiates these benchmarks from typical
commercial applications. 
CosmoFlow is trained on volumetric 3D data, rather than the 2D data commonly employed in training image classifiers.
DeepCAM is trained on images with $768 \times 1152$ pixels and 16 channels, which is substantially larger than standard vision datasets like ImageNet, where the average image is $469 \times 387$ pixels with at most 3 or 4 channels.
Moreover, the massive dataset sizes, 5.1 TB for CosmoFlow and 8.8 TB for DeepCAM, are over an order of magnitude larger than ImageNet (150GB) and introduce significant I/O challenges that expose storage and interconnect performance limitations. \changes{These characteristics offer unique performance differences from industrial machine learning workloads which must be addressed to optimize future systems where ML-augmented science applications are expected to play a critical role.} 
\section{Benchmarking Process}
\label{sec:benchmarking-process}
The MLPerf HPC benchmarking methodology is closely modeled after the MLPerf Training benchmark, including the general design, metrics, division rules, and submission and review procedures. For instance, the MLPerf HPC benchmarks use the same holistic and user-centric view of performance as MLPerf Training benchmarks and report \textit{time to train} as the primary metric. This choice creates a metric that is directly relevant to users and customers that captures end-to-end performance including both system speed and accuracy.
A few changes were made in the rules to improve the relevance of the benchmarks for scientific workloads in the HPC setting. For example, we introduced a rule to include data staging in the measured benchmark time to capture the impact of data-movement for massive scientific datasets on large HPC parallel file systems and node-local accelerated storage.

\subsection{Measurement}
Here we describe the finer details and rules relating to measuring \textit{time to train} performance in the benchmarks.

\subsubsection{Divisions}
MLPerf HPC has two types of submissions, \textit{closed} division and \textit{open} division. In the closed division, the submissions need to be equivalent to the reference implementation. This means that they must have mathematically equivalent model definitions and training algorithms. Such a process enables a direct comparison of the systems. In the open division, submitters are allowed to change the model architectures and training procedure freely but are restricted to evaluate the quality metric in the same way as the reference. This division aims to encourage innovations to further optimize the benchmarks.

\subsubsection{Timing rules}
At the start of a run, the benchmark dataset must reside on the parallel file system of the HPC center and on-node caches must be reset. We do not require resetting system-level caches because it is difficult to perform consistently across different systems and is often extremely disruptive. The benchmark timer begins as soon as the dataset is touched, which includes staging into node-local storage %\revision{(
\changes{such as an on-node SSD or RAM, if capacity is sufficient}.
The timer stops when the convergence criteria, as described in the rules, is met (Table \ref{tab:benchmarks-overview}).

\subsubsection{Run results}
\label{sec:run-results}
ML model training is inherently stochastic due to random initialization, dataset shuffling, etc. Therefore, to get an accurate measurement of the expected time to train, submitters must run the benchmarks a specified number of times to convergence. In the final scoring, we drop the fastest and slowest results and report the arithmetic mean of the remaining measured times.

\subsubsection{Logging}
The benchmarks use the \texttt{mlperf-logging} library~\cite{mlperf-logging}, which provides logging utilities and helper functions for all submissions. These help in collecting metadata and evaluating if the submissions meet compliance checks with the set run rules.

\subsection{Submission}

The submission process is designed to be fair, robust, and reproducible. This is achieved through the enforcement of a required structure for submissions and a peer review process. A submission schedule specifies when benchmarks and rules are finalized, when the submission window opens, the deadline for all submissions, as well as the schedule for the reviews and final deadline for results \changes{publication}. \changes{
A submission can be made to either open or closed divisions. A submitter can make as many submissions in either division and each submission entry can vary in the amount of system resources used.}

\subsubsection{Structure}
Reproducibility is a key goal for MLPerf HPC. Accordingly, submitters must upload their full code used to produce results, as well as system descriptions and the result log files containing the timing information. The submissions must conform to a specified file and directory structure and naming scheme for parsing, summarizing, and peer-review. The required submission structure is described in ~\cite{mlperf-submission-rules}.

\subsubsection{Review}

After the submission deadline, the peer-review process begins. A set of scripts from the \texttt{mlperf-logging} library are first used to check submissions and log files for compliance with the rules. Then, submitters review each other's implementations and results to further verify that they are compliant, sensible, and comprehensible. During the review stage, submitters are also allowed to perform ``hyperparameter borrowing'', in which they may perform additional sets of training runs using the hyperparameter settings of other submissions (but still using their original implementations). This discourages excessive hyperapameter tuning by submitters and avoids giving unfair advantages to teams with greater computational resources.

% System details table
\begin{table*}
\centering
%\footnotesize
\scriptsize
  \caption{HPC system details\\
  }
    \begin{threeparttable}
  \begin{tabular}{p{1.7cm}|r|l|p{1.5cm}|p{1.5cm}|p{2cm}|p{4cm}}
    \toprule
    \textbf{System} & \textbf{\#Nodes} & \textbf{Processors (per node)} & \textbf{Accelerators (per node)} & \textbf{Memory (per node)} & \textbf{Node-local storage (per node)} & \textbf{Interconnect topology and bandwidth} %\todo{clean this column}
    \\
    \midrule
    Piz Daint \cite{pizdaint} & 5,704 & 1x Intel Xeon E5-2690 v3      & 1x NVIDIA P100 (16 GB)            & 64 GB                                                        & N/A                  & Cray Aries (Dragonfly), 9.7 GB/s internode bi-directional  \\
    \hline
    ABCI \cite{abci}          & 1,088 & 2x Intel Xeon Gold 6148       & 4x NVIDIA V100 (16 GB)            & 384 GB                                                       & 1600 GB (SSD + NVMe) &  InfiniBand EDR (100Gbps) ×2, full-bisection bandwidth in the same rack (34 compute nodes) \\
       \hline
        %   Cori-KNL \cite{cori}      & 9,688 & 1x Intel Xeon Phi 7250        & N/A                               & \begin{tabular}{l} 96 GB DDR4 +\\ 16 GB MCDRAM \end{tabular} & N/A                  & Cray Aries (Dragonfly)
    Cori-KNL \cite{cori}      & 9,688 & 1x Intel Xeon Phi 7250        & N/A                               &  96 GB DDR4 + 16 GB MCDRAM  & N/A                  & Cray Aries (Dragonfly), >45 TB/s global peak bisection bandwidth      \\
       \hline
    Cori-GPU \cite{corigpu}   & 18 & 2x Intel Xeon Gold 6148          & 8x NVIDIA V100 (16 GB)             & 384 GB DDR4                                                  & 930 GB (NVMe)        & 4 dual-port Mellanox MT27800 ConnectX-5 EDR InfiniBand network (Fat Tree) \\
       \hline
    HAL \cite{ncsa-hal}       & 16 & 2x IBM POWER 9 model 2.2         & 4x NVIDIA V100 (16 GB)            & 256 GB DDR4                                                  & N/A                  & 2-Port EDR (Single Level) IB ConnectX-5 Adapter, 100 Gb/s  \\
       \hline
    Frontera-RTX \cite{tacc-frontera} & 90 & 2x Intel Xeon E5-2620 v4 &
    4x NVIDIA Quadro RTX 5000 (16 GB) 
    & 128 GB DDR4  & 240 GB (SSD)  & FDR InfiniBand MT27500 ConnectX-3 Adapter (Fat Tree), 56 Gb/s \\
    \hline
    Fugaku \cite{fugaku}      & 158,976 & 1x Fujitsu A64FX   & N/A & 32 GB \changes{HBM2} & 1.6 TB (NVMe SSD, shared among 16 compute nodes) & TofuD, (6D Mesh/Torus Network), 68GB/s x2 (in/out)  \\
    \hline
    ThetaGPU \cite{thetagpu} \tnote{*}
 & 24 & 2x AMD EPYC 7742  & 8x NVIDIA A100 (40 GB) & 1 TB DDR4  & 15TB SSD, 3.84TB NVMe      & 20 Mellanox QM9700 HDR200 40-port switches (Fat Tree), 25 GB/s node injection bandwidth \\
 \hline
    Summit  \cite{Summit} 
    \tnote{*}
 & 4,600 & 2x IBM 3.07 GHz POWER9        & 6x NVIDIA V100 (16 GB) & 512 GB DDR4   & 1.6TB (NVMe SSD)                   & 
 dual-rail EDR InfiniBand network (Fat Tree),  23GB/s node injection bandwidth 
 \\
    \bottomrule
  \end{tabular}
    \begin{tablenotes}
    \item[*] Measured performance metrics but did not submit for v0.7 submissions
    \end{tablenotes}
    \end{threeparttable}
    % }
    \vspace{-.10in}
      \label{tab:systemdetails}
\end{table*}

\section{Results}
\label{sec:results}
The inaugural MLPerf HPC submission round (v0.7)\footnote[2]{naming chosen to be consistent with submission round in MLPerf Training} took place during the summer of 2020. The results  from submissions on 7 supercomputers, released in November,

showcased the capabilities of HPC systems listed in Table~\ref{tab:systemdetails}, 
for training large-scale scientific problems.
The results are summarized in Table~\ref{tab:closed-results} for both closed and open divisions. 

\mlhpc{At first glance, we observe a system scale range of up to a factor of 8-16$\times$ %for both CPU- and GPU-based systems 
in closed division for both benchmarks and results showing $> 10 \times$ improvements in time to train across this range.
Furthermore, we can see the effect of innovations in open division by comparing each of these submissions to the fastest submission on the same system in closed division. For instance, on the CPU-based system Fugaku, there is an 3.38$\times$ improvement in open division for CosmoFlow while for the GPU-based system ABCI, there are speedups of 2.61$\times$ and 1.12$\times$ in open division for CosmoFlow and DeepCAM respectively.\footnote{\changes{These numbers are comparing time to solution in open and closed division submissions from the same system, which does not necessitate equal number of processors/accelerators in both submissions (cf. Table \ref{tab:closed-results}).}}}

\mlhpc{In this section}, we present a detailed analysis of the submissions \changes{in a framework that allows to clearly understand and separate the effects of data-staging, algorithmic convergence and system compute throughput (training and evaluation) in time to solution from the submitted logs} \mlhpc{(subsection \ref{ssec:results-analysis})} \changes{and complement this with a set of highlights on implementations from few systems} \mlhpc{(subsection \ref{ssec:highlights})}.

\begin{table*}
\centering
\footnotesize
  \caption{%Closed division results, reporting the
  Performance metrics (time to solution in minutes) from submissions in closed and open divisions }
  \label{tab:closed-results}

  \begin{threeparttable}
  \begin{tabular}{lllp{2.2cm}rrrrr}
    \toprule
  \textbf{Division} &  \textbf{System} & \ \ \ \textbf{Submission} & \textbf{Software} & \textbf{\#Processors} & \textbf{\#Accelerators} & \mlhpc{
  \textbf{Parallelism}\tnote{\textdagger}
  } & \textbf{CosmoFlow} & \textbf{DeepCAM} \\
    \midrule
    
   Closed &  Piz Daint    & \ \ Piz-Daint-128   & TensorFlow 2.2.0                     & 128    & 128   &  2 s/1 GPU\ \ \  & 461.01 & -	     \\
          &  Piz Daint    & \ \ Piz-Daint-256   & TensorFlow 2.2.0                     & 256    & 256   &  2 s/1 GPU\ \ \  & 327.01 & -	     \\
          &  ABCI         & \ \ ABCI-1024       & PyTorch 1.6.0                        & 512    & 1,024 &  2 s/1 GPU\ \ \  & -      & 11.71  \\
          &  ABCI         & \ \ ABCI-512        & TensorFlow 2.2.0                     & 256    & 512   &  1 s/1 GPU\ \ \  &  34.42 & -	     \\
          &  Fugaku       & \ \ Fugaku-512      & TensorFlow 2.2.0 + Mesh TensorFlow   & 512    & -     &  1 s/1 CPU\ \ \  & 268.77 & -	     \\
          &  Fugaku       & \ \ Fugaku-8192     & TensorFlow 2.2.0 + Mesh TensorFlow   & 8,192  & -     &  1 s/16 CPUs\ \  & 101.49 & -	     \\
          &  Cori-GPU     & \ \ Cori-GPU-64     & PyTorch 1.6.0                        & 16     & 64    &  2 s/1 GPU\ \ \  & -      & 139.29 \\
          &  Cori-GPU     & \ \ Cori-GPU-64     & TensorFlow 1.15.0                    & 16     & 64    &  1 s/1 GPU\ \ \  & 364.73 & -      \\
          &  Cori-KNL     & \ \ Cori-KNL-512    & TensorFlow 1.15.2                    & 512    & -     &  1 s/1 CPU\ \ \  & 536.06 & -      \\
          &  HAL          & \ \ HAL-64          & TensorFlow 1.15.0                    & 32     & 64    &  1 s/1 GPU\ \ \  & 265.59 & -      \\
          &  Frontera-RTX & \ \ Frontera-RTX-64 & TensorFlow 1.15.2                    & 32     & 64    &  1 s/1 GPU\ \ \  & 602.23 & -      \\
  \hline                                                                                                              
    Open & ABCI        & $\star$ABCI-1024       & PyTorch 1.6.0                        & 512    & 1,024 & 2 s/1 GPU\ \ \   & -      & 10.49  \\
         & ABCI        & $\star$ABCI-2048       & TensorFlow 2.2.0                     & 1,024  & 2,048 & 1 s/1 GPU\ \ \   & 13.21  & -      \\
         & Fugaku      & $\star$Fugaku-16384    & TensorFlow 2.2.0 + Mesh TensorFlow   & 16,384 & -     & 1 s/4 CPUs\ \    & 30.07  & -      \\
         & Cori-KNL    & $\star$Cori-KNL-1024   & TensorFlow 1.15.2                    & 1,024  & -     & 1 s/1 CPU\ \ \   & 419.69 & -      \\  

\bottomrule
\end{tabular}
    \begin{tablenotes}
    \item[\textdagger] \mlhpc{Data-parallel granularity of train step: \# samples (s) processed by number of compute units forming a data-parallel unit in each train step. E.g. Piz-Daint-128 processes $2$ samples ("local batch size") on each GPU (pure data-parallelism, batch size $128 \times 2 = 256$), whereas Fugaku-8192 processes $1$ sample in each group of $16$ CPUs (through model-parallelism within this group, data-parallelism across these groups of which there are $8192/16 = 512 =$ batch size).

    }
    \end{tablenotes}
  \end{threeparttable}
%   }
\vspace{-.10in}
\end{table*}

\begin{figure}[!b]
\vspace{-.1in}
  \centering
  \includegraphics[width=\linewidth]{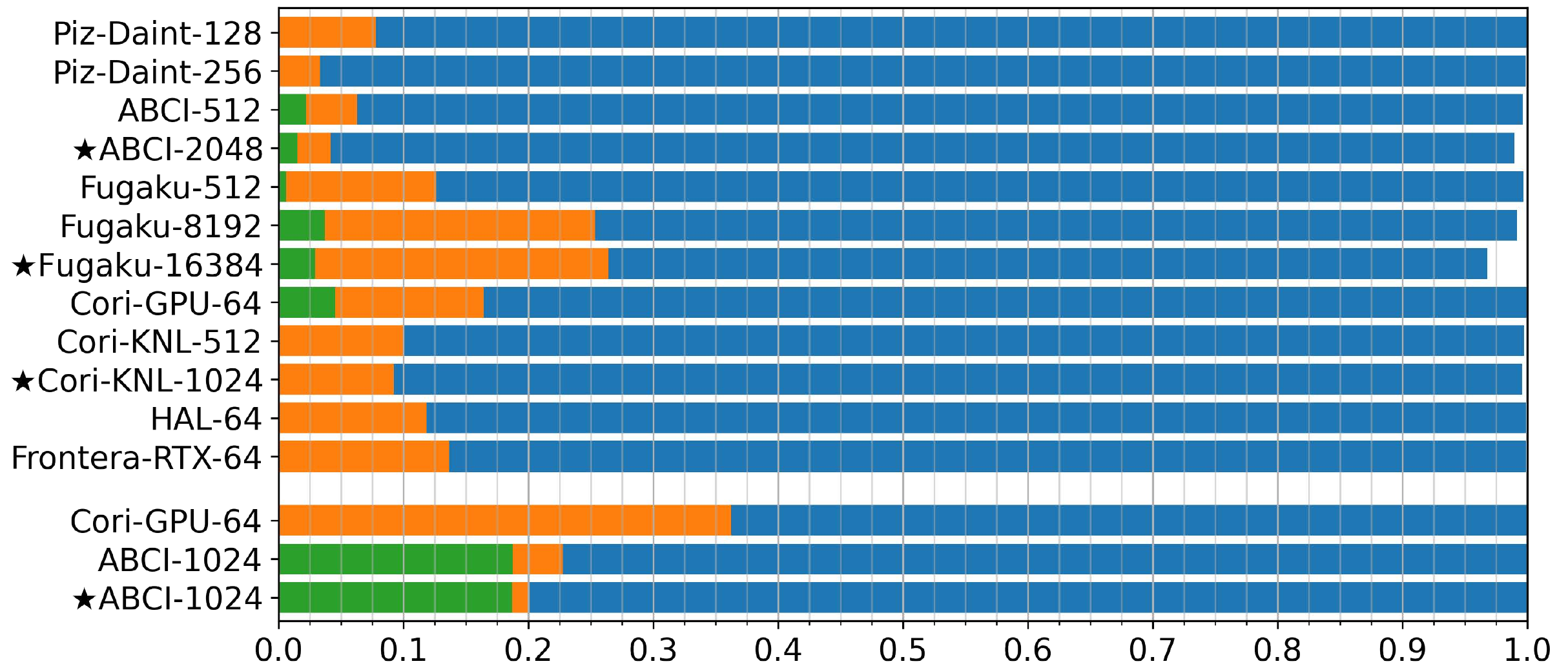}
  \vspace{-4mm}
  \caption{Relative breakdown of time to train normalized to range [0-1], into staging (green), evaluation (orange) and training (blue). Lower three entries on y-axis are for DeepCAM, rest are for CosmoFlow.} 
  \label{fig:analysis-times-relative-breakdown}
  \vspace{-.2in}
\end{figure}

\begin{table}[!b]
\vspace{-.2in}
\footnotesize
  \caption{Data staging time}
    \centering
    
\resizebox{0.90\columnwidth}{!}
{    
  \begin{tabular}{cccc}
        \toprule
        \textbf{Benchmark} & \textbf{Submission} & 
        \begin{tabular}{c}\textbf{Staging time} \\ (\textbf{minutes}) \end{tabular}
        & 
        $\frac{T_{staging}}{T_{epoch}}$ \\
        \midrule
CosmoFlow & Cori-GPU-64             &               $16.49 \pm 0.61 $ & 2.55 \\
          & \ ABCI-512\             &                $0.76 \pm 0.004$ & 2.27 \\
          & $\star$ABCI-2048        &                $0.20 \pm 0.004$ & 1.56 \\
          & \ Fugaku-512\ \         &                $1.55 \pm 0.11 $ & 0.64 \\
          & \ Fugaku-8192\          &                $3.77 \pm 0.51 $ & 3.59 \\
          & $\star$Fugaku-16384     &                $0.88 \pm 0.08 $ & 4.93 \\
\midrule
DeepCAM   & \ ABCI-1024\            &                $2.20 \pm 0.01$  & 5.55 \\
          & $\star$ABCI-1024        &                $1.96 \pm 0.08$  & 5.45 \\
        \bottomrule
    \end{tabular}
    }
    \label{tab:staging-time}
    \vspace{-5mm}
\end{table}

\subsection{Analysis}
\label{ssec:results-analysis}
\noindent The time to solution broken into data staging, training and evaluation components is shown in Fig. \ref{fig:analysis-times-relative-breakdown} for each submission.

\subsubsection{Discussion of data staging time}
\label{sec:data-staging}

The data staging time, $T_{staging}$, is shown in Table \ref{tab:staging-time} for the systems where it was measured. 
We observe that it is very different for the two benchmarks on ABCI - by interpolation to 1,024 GPUs, staging \changes{is handled} more than $5 \times$ faster \mlhpc{(in absolute time)} \changes{for CosmoFlow} than for DeepCAM. 
This difference comes not only from the fact that DeepCAM's data set is 73\% larger than CosmoFlow's,  but also from the data \changes{compression ratio, which is 88\% for CosmoFlow compared to only 23 \% for DeepCAM}\footnote{\changes{As a consequence, compression was not applied in data staging for DeepCAM.}}.
%\Lukas{@Koichi: Can it be explained by some optimizations? E.g. different compression in CosmoFlow vs DeepCAM or similar?} 

To understand the relative importance of staging ($T_{staging}$) in time to solution, $T_{solution}$, we assume that % the runtime model 
\begin{equation}
T_{solution} = T_{staging} + T_{compute} + T_{extra}
\label{eq:time-to-solution}
\end{equation}
With $T_{compute} = T_{epoch} \cdot \# epochs$ and $T_{extra} \approx 0$ (Fig. \ref{fig:analysis-times-relative-breakdown}), where $T_{epoch}$ is the average epoch time, we get 
\begin{equation}
\frac{T_{staging}}{T_{solution}} \approx \frac{T_{staging}/T_{epoch}}{T_{staging}/T_{epoch} + \# epochs}
\label{eq:relative-staging-time}
\end{equation} 
The ratio $T_{staging}/T_{epoch}$, thus, quantifies the \textit{relative} overhead in units of compute epochs that staging adds to time to solution irrespective of convergence\footnote{This number depends on the batch size through $T_{epoch}$, though.} %} through $T_{epoch}$.} %, though.}.
\changes{or the exact amount of data used}\footnote{\mlhpc{As long as it does not traverse a capacity limit in the memory hierarchy closer to compute units than the staging target (e.g. RAM capacity if staging to on-node SSD).}}. 
Since it relates the compute to staging throughput, it is, however, dependent on the \changes{model's computational complexity}. As DeepCAM is more compute-intensive than CosmoFlow, the relative overhead in "extra compute epochs" shown in the last column of Table \ref{tab:staging-time} is reduced to a factor of $2.5-3.5 \times$ that of CosmoFlow from what is expected purely from dataset compressibility. The reason for DeepCAM on ABCI still having $10 \times$ the share of staging compared to CosmoFlow, is that CosmoFlow requires $4 \times$ more epochs to converge (Table \ref{tab:deepcam-vs-cosmoflow-results}), which causes all its submissions to have marginal share of staging ($<$ 5\% in Figure \ref{fig:analysis-times-relative-breakdown}).

Finally, Fugaku's staging times are the highest at 8,192 processors \changes{due to} 16-way replication of the data set (4-way for 16,384). This is an overhead of model-parallelism and could be avoided by \changes{partitioned} reading and broadcasting of data across MPI ranks \mlhpc{within each model-parallel group}.

These findings show that the overhead of staging is \mlhpc{highly application-specific}, \mlhpc{depending} on data compressibility, but also a model's \changes{computational complexity} and convergence % of an application, 
and generally affects smaller systems with fewer epochs to converge more than larger ones.

\begin{figure*}[t]
  \centering
  \includegraphics[width=\textwidth]{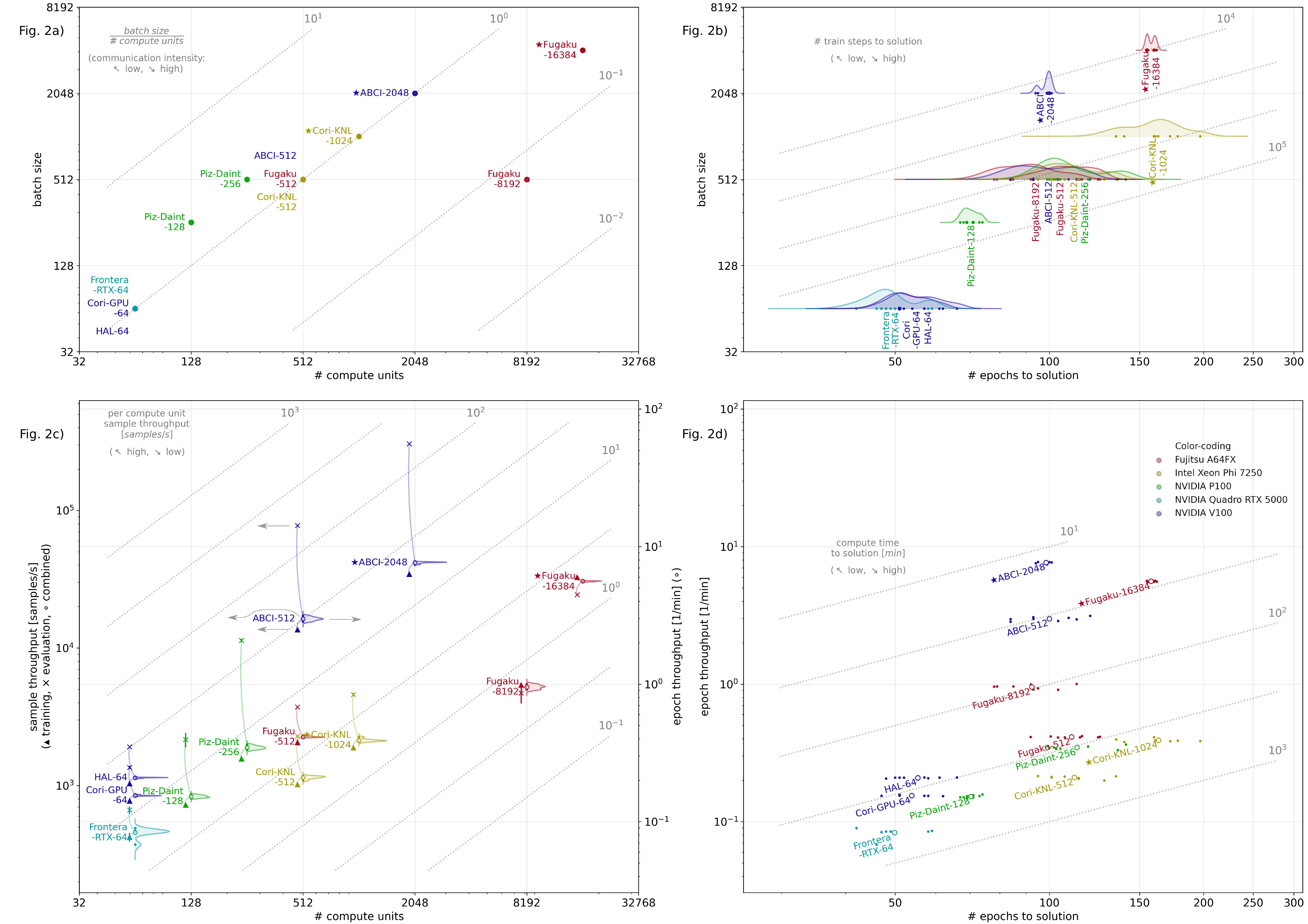}
  \vspace{-6mm}
  \caption{Compute analysis of CosmoFlow with figures on a) parameter choice, b) epoch scaling (\textit{dependent variable on x-axis}), c) throughput scaling (arrows %in $\star$Fugaku-16384 explain how to read x-axis, 
  in ABCI-512 explain how to read y-axes, distribution is given for the combined/epoch throughput, curved lines illustrate averaging of training and evaluation to combined/epoch throughput), and d) compute time. Note that axes are \textit{shared} along rows (y-axis in a \& b: \textit{batch size}, in c \& d: \textit{epoch throughput}) and columns (x-axis in a \& c: \textit{\# compute units}, in b \& d: \textit{\# epochs to solution}).}
  \label{fig:analysis-cosmoflow-benchmark-overview}
  \vspace{-.25in}
\end{figure*}

\subsubsection{Analysis of compute time}
\label{sec:compute-time-analysis}

This subsection presents an analysis of the time spent in training and evaluation, $T_{compute}$ in eq. \eqref{eq:time-to-solution}, after data staging is completed. It represents $\geq 90\%$ of time to solution in CosmoFlow and $\geq 80\%$ in DeepCAM (Figure \ref{fig:analysis-times-relative-breakdown}). 

\textit{(a) Compute analysis for CosmoFlow:}
We observe that the number of epochs to convergence in $T_{compute} = T_{epoch} \cdot \text{\#epochs}$  is primarily an algorithmic property of SGD \mlhpc{(holding the data and model fixed)}, and as such, only dependent on the optimizer and choice of hyperparameters, but not the particular system or implementation (up to floating point precision).
On the other hand, $T_{epoch}$ is a system-specific property that depends on both the hardware and the specific parallel implementation of the model as well as the choice of optimizer, but not necessarily on all hyperparameters. In fact, from the rule set in Table \ref{tab:benchmarks-overview}, $T_{epoch}$ only depends on the value of the \textit{batch size} hyperparameter.
Therefore, we provide an analysis of the epoch throughput $T_{epoch}^{-1}$ and number of epochs to converge separately for each of the submissions by means of Figure \ref{fig:analysis-cosmoflow-benchmark-overview}. 

\changes{\textit{Submission parameters:}}
We focus on the number and type of compute unit (accelerators for GPU-based systems, processors for CPU-based systems) to characterize the system and the batch size as most important parameters. The choice of these parameters for each submission is shown in Figure \ref{fig:analysis-cosmoflow-benchmark-overview} (a).
\mlhpc{
The ratio of batch size to number of compute units gives the amount of samples processed by a single compute unit on average per train step. Together with the parallelism column in Table \ref{tab:closed-results} that characterizes the data-parallel unit per train step, we can observe that all CosmoFlow submissions with a ratio $\geq 1$ chose a purely data-parallel implementation - Piz Daint being the only with a local batch size of 2, all others 1 - whereas those with a ratio $< 1$, i.e. Fugaku-8192 and $\star$Fugaku-16384, used model-parallelism in addition within each data-parallel unit of 16 and 4 CPUs respectively, resulting in a hybrid form of model- and data-parallelism that will be discussed in subsection \ref{sssec:fugaku}.
We note that submissions with the same batch size to number of compute units ratio (constant along diagonal lines) that use the same data-parallel unit (Table \ref{tab:closed-results}) are related by (data-parallel) weak scaling. From another point of view, this ratio roughly behaves inversely proportional to the communication intensity, i.e. the amount of communication per computation, per train step for a given type of parallelism\footnote{assuming communication to be roughly constant per train step with data-parallel scale-out and the amount of computation scaling with the batch size}, so that regions of different communication intensity can be identified in Figure \ref{fig:analysis-cosmoflow-benchmark-overview} (a).
}

\changes{\textit{Epoch scaling:}}
In Figure \ref{fig:analysis-cosmoflow-benchmark-overview} (b), we show the scaling of epochs required to converge as a function of the batch size (dependent variable on the x-axis). As discussed above, this is a \textit{system-independent} property up to floating point precision. The level lines along the diagonal identify points of an identical number of train steps to solution, which puts a limit on data-parallel scalability. That is, once an increase in the batch size leads to a larger number of train steps to solution, a system can no longer train a model faster by growing the compute resources proportionally (ignoring caching effects in the memory hierarchy). The submissions $\star$ABCI-2048 and $\star$Fugaku-16384 are close to this limit (cf. Figure \ref{figure:fugaku-epoch-time}) and the specialized techniques to converge at these very large batch sizes will be discussed in greater detail in subsection \ref{sssec:fugaku}.
The remaining submissions all closely follow the reference implementation \mlhpc{described in subsection \ref{ssec:benchmark-suite-cosmoflow}}.
This turns out to scale efficiently to a batch size of 256 with only $1.3\times$ more epochs to converge for $4\times$ batch size increase from 64, but past this point becomes significantly harder to train and less stable ($1.6 \times$ epochs for doubling the batch size). \changes{Closed division submissions were limited to a maximum batch size of 512 due to convergence issues at batch size 1,024 that were only overcome by $\star$Cori-KNL-1024 in open division with a slight modification of the learning rate schedule (section \ref{sssec:cori})
}.

\changes{\textit{Throughput scaling:}}
\mlhpc{
In Figure \ref{fig:analysis-cosmoflow-benchmark-overview} (c), compute throughput is shown as a function of the number of compute units and, implicitly, the batch size through the shared x-axis with Figure \ref{fig:analysis-cosmoflow-benchmark-overview} (a). 
For each submission, we plot sample throughput (left axis) for training ($\blacktriangle$) and evaluation ($\times$) as well as the combined sample throughput (\# samples/$T_{epoch}$, distribution with mean at $\circ$\footnote{The curved lines from training ($\blacktriangle$) and evaluation ($\times$) rooted at the combined mean ($\circ$) illustrate the averaging by holding the point cloud together.}) according to the split of the data set (80\% training and 20\% evaluation) at the abscissa corresponding to the number of compute units (illustrated on the ABCI-512). 
On the diagonal we find lines of constant per-compute-unit throughput, commonly used to analyze scaling efficiency. Dividing the combined throughput by the overall number of samples in the data set, we obtain the epoch throughput, $T_{epoch}^{-1}$, which can be read off for the distribution from the right scale. 
Note that throughput is only algorithmically relevant together with a specified batch size, so that Figure \ref{fig:analysis-cosmoflow-benchmark-overview} (c) and (a) are meant to be read together.}
The resulting \mlhpc{pair of plots} allows us to understand the scaling efficiency of training, evaluation, and combined throughput for submissions that relate through weak \mlhpc{(same data-parallel unit in Table \ref{tab:closed-results} and same diagonal in Figure \ref{fig:analysis-cosmoflow-benchmark-overview} (a))} or strong \mlhpc{(same batch size in Figure \ref{fig:analysis-cosmoflow-benchmark-overview} (a))} scaling, or extrapolation thereof, and compare different systems with each other. 

Interestingly, \mlhpc{we observe that} for GPU-based systems, there is a transition occurring from smaller systems to those with $256$ GPUs and more, where the gap between training and evaluation throughput becomes very high. Evaluation is inherently bound by data access speed in CosmoFlow and it turns out that the memory configurations of these systems is exactly such that the larger ones, i.e. Piz-Daint-256, ABCI-512 and $\star$ABCI-2048, benefit from caching the dataset in RAM.
As a result, these systems spend very little ($<$ 5\%) of their time in evaluation (Figure \ref{fig:analysis-times-relative-breakdown}). 
\changes{This is different for the CPU-based systems that are used at larger processor count and, thus, higher RAM capacity. To reduce the I/O-related performance penalty, smaller systems could selectively cache the evaluation dataset in RAM.}

\changes{
At the other end of the scale, we can observe the network effects on training throughput. Specifically, the difference in scaling efficiency between training and evaluation for submissions related by weak-scaling shows that training at scale increasingly becomes network-bound whereas evaluation with its low communication overhead almost scales ideally. This is the case for e.g. ABCI-512 and $\star$ABCI-2048 - the system with the highest measured throughput in this round - with 63.2 \% efficiency for training vs. 98.2 \% for evaluation and similarly CoriKNL-512 and $\star$CoriKNL-1024 (92.6 \% for training vs. 100.3 \% for evaluation).
}

\changes{A further interesting point is the relation of HAL-64, CoriGPU-64 and ABCI-512 that all have the same accelerator type but a different CPU-to-GPU ratio. Without the optimizations on ABCI described in \ref{sssec:abci}, its per-GPU training performance is similar to HAL-64. 
Both of these systems have one CPU per 2 GPUs and the same number of GPUs per node. In contrast, Cori-GPU represents a more consolidated system with 8 GPUs per node, which reduces the load on the network, but also with smaller CPU resources per GPU at one CPU per four GPUs. This difference can play a role in implementations with significant CPU-overhead to keep the GPUs busy and here in parts explains the uniform difference seen in training and evaluation throughput between HAL and Cori-GPU.}

\begin{figure}[htbp]
\vspace{-.1in}
  \centering
  \includegraphics[width=0.95\linewidth]{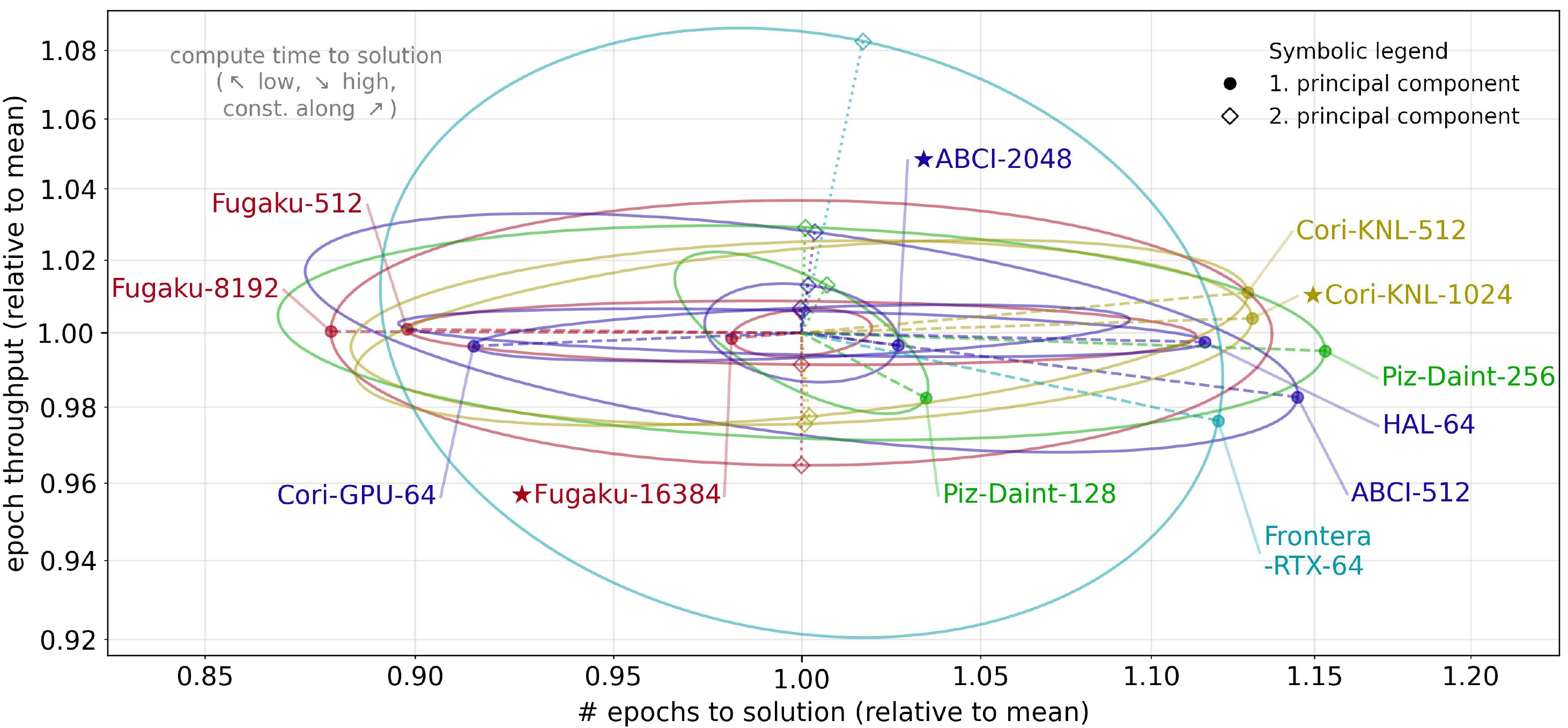}
  \vspace{-3mm}
  \caption{\mlhpc{Log-PCA on epochs \& throughput (relative scales) in CosmoFlow. Note that $\log ( T_{\text{compute}}) = \log(\text{\# epochs}) - \log(\text{epoch throughput})$. Half-axes of ellipses correspond to standard deviation of principal components. }} 
  \label{fig:analysis-log-pca-epochs-throughput}
  \vspace{-.1in}
\end{figure}

\changes{\textit{Compute time:}}
Figure \ref{fig:analysis-cosmoflow-benchmark-overview}(d) shows $T_{compute}$ - the outcome of the competition between epoch and throughput scaling when growing the system. To obtain it, we join the epoch scaling (Figure \ref{fig:analysis-cosmoflow-benchmark-overview}b) and throughput scaling (Figure \ref{fig:analysis-cosmoflow-benchmark-overview}c/a) of each submission \textit{on the batch size} along the shared axes. The compute time $T_{compute} = T_{epoch} \cdot \# \text{epochs}$ is constant along diagonal lines indicated. We observe that while Fugaku-512 has a higher sample throughput than HAL-64, the smaller batch size of HAL-64 causes it to still converge slightly faster in time overall \changes{(similarly for $\star$Cori-KNL-1024 and Piz-Daint-256)}. 
As a further insight, we are able to trace back the run-to-run variation in time to solution, %(Table \ref{tab:closed-results})
\changes{which is 2.8\% for $\star$ABCI-2048, $\star$Fugaku-16384 and 11.1\% for all other CosmoFlow submissions,} to the number of epochs rather than system throughput. 
\mlhpc{This is confirmed by a logarithmic principle component analysis (log-PCA, Figure \ref{fig:analysis-log-pca-epochs-throughput}), which shows a first component with strong horizontal alignment in all submissions. Furthermore, we see that Frontera-RTX-64 exhibits especially high relative throughput variability, which is explained by the parallel Lustre filesystem from where data is loaded continuously during training and that is shared by thousands of users at any time.}
Finally, the technique of grouping throughput- and epoch-scaling plots together in this setting allows the inexpensive prediction of compute time to convergence at system configurations other than the ones used in the submissions. 
This can be done either by additional throughput measurements in Figure \ref{fig:analysis-cosmoflow-benchmark-overview}(c) at batch size with known epochs to convergence or approximately\footnote{\changes{Ignoring network effects.}} by data-parallel extrapolation
of throughput along diagonal lines of constant throughput per compute unit\footnote{\changes{Weak scaling: batch size/\# compute units = const., ignoring effects of dataset size}} and joining that value with the known epoch scaling from Figure \ref{fig:analysis-cosmoflow-benchmark-overview}(b) on the batch size.
Further details on CosmoFlow's \mlhpc{results} are presented in section~\ref{ssec:highlights}.

\textit{(b) Compute analysis for DeepCAM:}
Figure \ref{fig:analysis-times-relative-breakdown} shows that more than a third (36.3\%) of time to solution is spent in the evaluation phase on Cori-GPU compared to \changes{only 4\% (closed) and 1 \% (open) on ABCI}. The reason for this is not primarily the different evaluation throughput (Table \ref{tab:deepcam-vs-cosmoflow-results}), but that evaluation is triggered after a fixed number of training steps instead of once per epoch and delayed in open division. As a consequence, Cori-GPU calculates the IOU score $8 \times$ and $35 \times$ more often than ABCI's closed and open submissions respectively.
We omit \changes{a graphical analysis of DeepCAM similar to CosmoFlow} here due to the lack of space \mlhpc{(it is available from the linked artifact in the appendix)}.

\textit{(c) Compute analysis comparison:} 
Table \ref{tab:deepcam-vs-cosmoflow-results} summarises the compute characteristics of both applications in the benchmark suite on Cori-GPU and ABCI.  
\changes{
DeepCAM requires only 20-25\% the number of epochs to converge of CosmoFlow,
but the epochs growth rate as a function of the batch size is similar for both applications. DeepCAM, however, exhibits much more stable convergence (1.7\% per-run std deviation of time to solution vs. 11.1\% for CosmoFlow submissions following the reference implementation).} 
For sample throughput, we find that (1) CosmoFlow has higher throughput in both training ($3-5 \times$) and evaluation ($7-20 \times$), which can be attributed to the lower number of layers, whereas (2) DeepCAM has a much smaller gap between training and evaluation throughput ($1.4\times$ vs. $5.7\times$).
Comparing the resource footprint, we find that to reach convergence, the compute budget (time to solution in hours  $\times$ number of accelerators/processors) is $1.5-2.6 \times$ larger for CosmoFlow than for DeepCAM, with correspondingly higher time to solution that is $2.6-2.9 \times$ that of DeepCAM for closed division. 
Notably, though, we see for both benchmarks that despite a relatively large increase in number of accelerators by $8-16 \times$, the compute budget with the right optimizations can be \mlhpc{controlled} and a decrease in $\sim 10 \times$ time to solution is reliably possible.

\begin{table*}
\small
\caption{Scaling of required epochs, throughput, time to solution and compute budget in CosmoFlow vs. DeepCAM.}
\centering
\resizebox{1.99\columnwidth}{!}
{
\begin{tabular}{ccccccccc}
\toprule
\textbf{Benchmark} & \textbf{Submission} &  \begin{tabular}{c}\textbf{Batch} \\ \textbf{size} \end{tabular} &  \textbf{\# Epochs} & \textbf{\# GPUs} & \begin{tabular}{c}\textbf{Training} \\ \textbf{throughput/\# acc.} \\ (samples/second) \end{tabular} & \begin{tabular}{c}\textbf{Evaluation} \\ \textbf{throughput/\# acc.} \\ (samples/second) \end{tabular} & \begin{tabular}{c}\textbf{Time to solution} \\ (minutes) \end{tabular} & \begin{tabular}{c}\textbf{Compute}\\ \textbf{budget} \\ 
(h $\cdot$ acc) 
\end{tabular} \\
\midrule
CosmoFlow & Cori-GPU-64             &     64 &    $53.88 \pm  4.85$  &   64  & $12.07 \pm 0.09$  &  $21.17 \pm 0.56$ & $364.73 \pm 32.77$ & 389.04 \\
          & \ ABCI-512\             &    512 &   $100.00 \pm 13.50$  &  512  & $26.59 \pm 0.90$  & $151.88 \pm 0.68$ & $34.42 \pm 4.03$ & 293.03 \\
          & $\star$ABCI-2048        &  2,048 &    $98.50 \pm  2.56$  & 2,048 & $16.79 \pm 0.23$  & $149.21 \pm 0.28$ & $13.21 \pm 0.35$ & 450.96 \\
\midrule
DeepCAM   & Cori-GPU-64             &   128  &    $10.00 \pm  0.00$ &   64  & $ 3.56 \pm 0.13$ &   $3.55 \pm 0.18$ & $139.29 \pm 3.63$ & 148.58 \\
          & \ ABCI-1024\            &  2,048 &    $24.00 \pm  0.00$ & 1,024 & $ 5.24 \pm 0.02$ &   $7.37 \pm 0.01$ & $11.71 \pm 0.02$ & 199.78 \\
          & $\star$ABCI-1024        &  2,048 &    $23.67 \pm  1.16$ & 1,024 & $ 5.57 \pm 0.13$ &   $4.67 \pm 0.82$ & $10.49 \pm 0.23$ & 178.95 \\
\bottomrule
\end{tabular}
}
\label{tab:deepcam-vs-cosmoflow-results}
\vspace{-.2in}
\end{table*}

\subsection{Highlights}
\label{ssec:highlights}
In this section, we present details of the implementations and present highlights from v0.7 submissions.

\subsubsection{Fugaku}
\label{sssec:fugaku}

\begin{figure}[htbp]
    \subfigure[]{\includegraphics[width=0.48\linewidth]{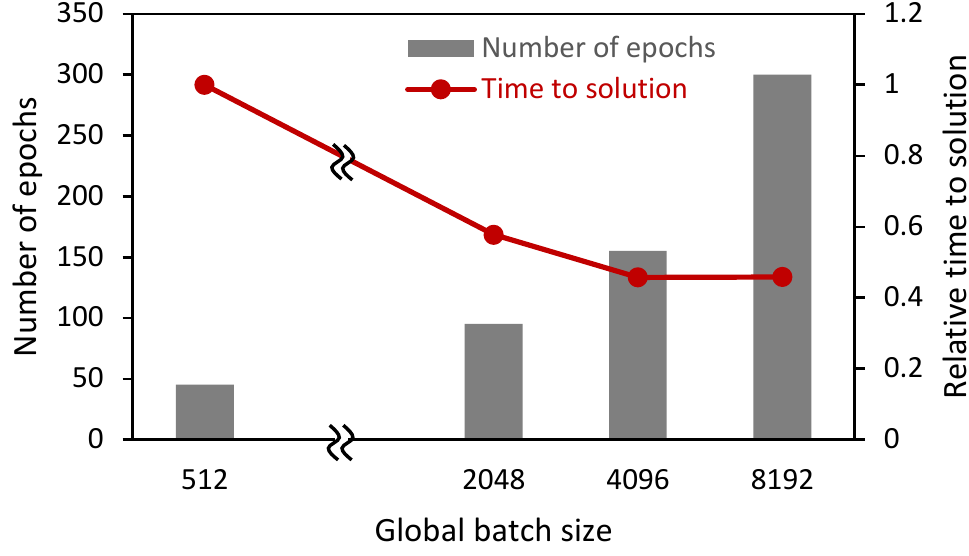}
    \label{figure:fugaku-epoch-time}}
    \subfigure[]{\includegraphics[width=0.48\linewidth]{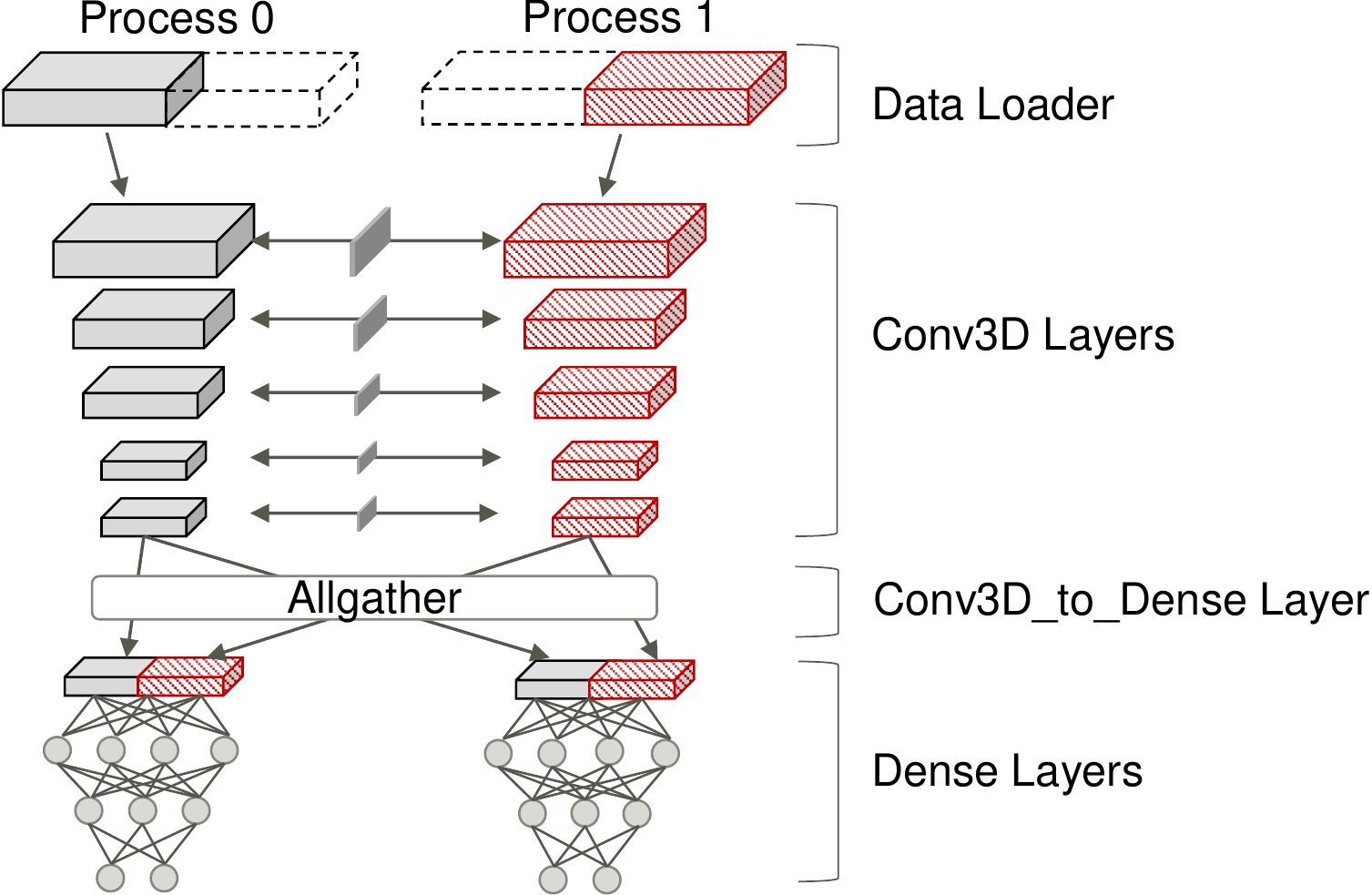}
    \label{figure:model-parallelism}}
    \vspace{-2mm}
     \caption{\changes{(a) The number of epochs to solution (left y-axis) in open division for CosmoFlow (see $\star$ABCI-2048, $\star$Fugaku-16384 in Figure \ref{fig:analysis-cosmoflow-benchmark-overview}b) and extrapolated time to solution (right y-axis, relative to 512 CPUs) as a function of global batch size on Fugaku with data parallelism (local batch size is one). (b) Spatial partitioning (2x1) for \changes{model-parallelism in} CosmoFlow.
    }
     }\end{figure}

\begin{figure}[htbp]
    % \vspace{-.2in}
   % \centering
    \subfigure[Using data parallelism]{\includegraphics[width=0.48\linewidth]{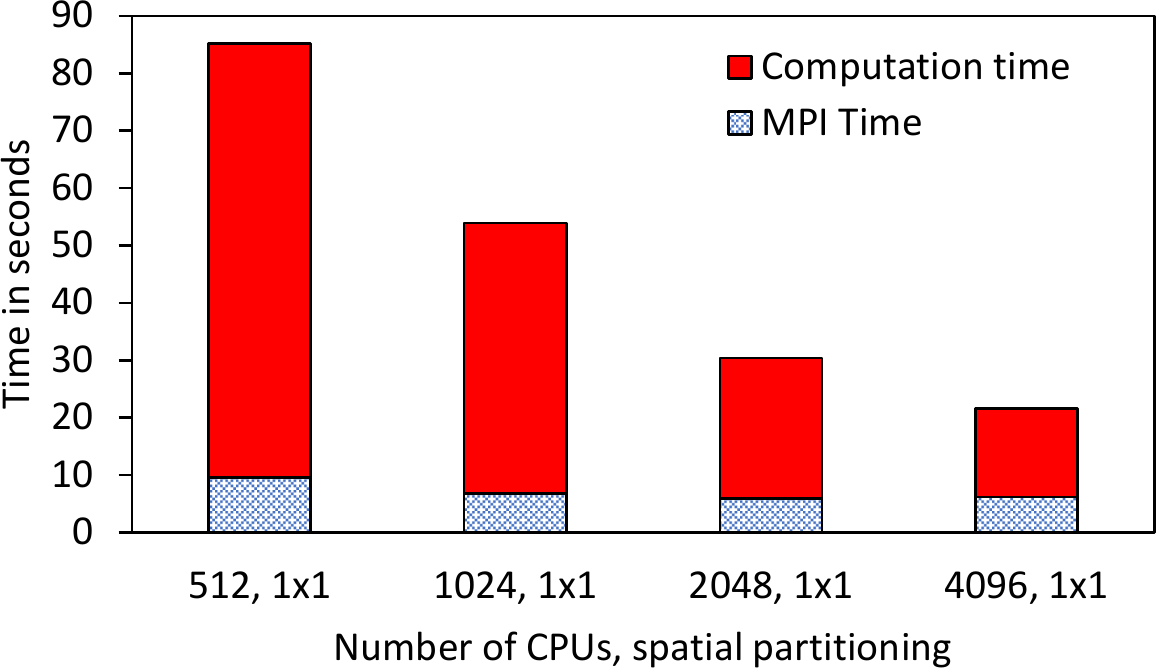}}
    \label{sfigure:fugaku-time-2epochs-data-parallel}
    \subfigure[Using model parallelism]{\includegraphics[width=0.48\linewidth]{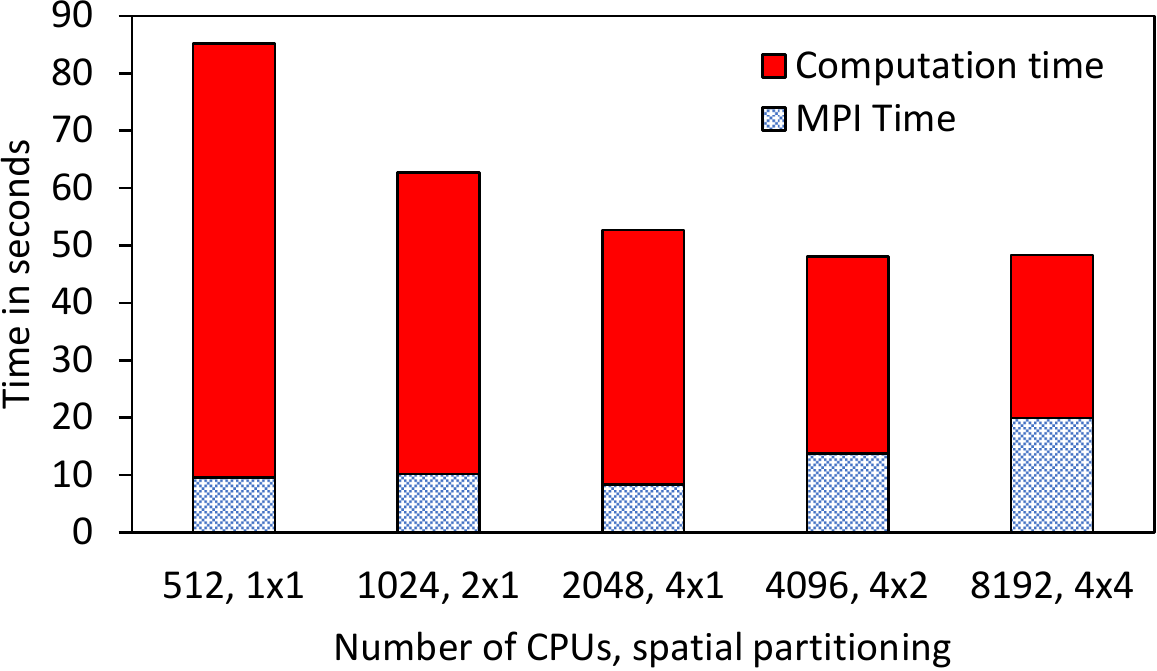}
    \label{sfigure:fugaku-time-2epochs-model-parallel}}
    \vspace{-2mm}
     \caption{\changes{Elapsed time for the first two epochs of CosmoFlow on Fugaku \changes{using data-parallel scaling (local batch size one, no spatial partitioning) and (hybrid) model-parallel scaling (global batch size 512, with spatial partitioning) starting from a data-parallel baseline with 512 CPUs. Time spent in communication (MPI) is measured as in section \ref{sec:workload-characterization}.}}
     }
    \label{figure:fugaku-time-2epochs}
    \vspace{-.2in}
\end{figure}

Submissions on the Fugaku supercomputer to CosmoFlow
utilized hybrid data- and model-parallel execution.
\changes{To reduce staging time,} data reformatting by compressing and archiving multiple files was effective.
Training data is staged in RAM disks in advance. Only in the case of the 512-node submission, RAM disks do not have enough capacity, so data staging is performed to local SSDs on I/O nodes that are assigned to each unit of 16 nodes. Also, the data cache function of TensorFlow (\texttt{tf.data.Dataset.cache}) is used to improve the bandwidth of data loading during training.

For the open division, the following accuracy improvement techniques were applied: (1) use linear learning rate decay schedule, (2) apply data augmentation, (3) disable dropout layers.
\changes{
When applying the three techniques in isolation at batch size 1,024, the number of epochs until reaching a relaxed target of MAE $<$ 0.15 (cf. Table \ref{tab:benchmarks-overview})
is (1) 64, (2) 68, (3) 19, compared to 67 without any of the three. Although disabling dropout layers is the most effective among the three, none of them converges to the target MAE $<$ 0.124 alone.
Applying all the three techniques together is necessary to increase the batch size above the closed division bottleneck of 512. 
The resulting epoch scaling for the open division is shown in Figure~\ref{figure:fugaku-epoch-time} together with extrapolated time to solution for data-parallelism on Fugaku (based on the submission at 512 CPUs and ideal throughput scaling\footnote{Using $T_{solution} \propto T_{staging}/T_{epoch} + \#\text{epochs}$ from equation \ref{eq:time-to-solution}.}). We see that the time to solution ideally scales up to batch size 4,096 with data-parallelism at local batch size one, which was confirmed experimentally on Fugaku.
To summarize, the result of open division optimizations on the CosmoFlow submissions can be seen in Figure \ref{fig:analysis-cosmoflow-benchmark-overview}b - the batch size can be increased from the closed division bottleneck at 512 to 2,048 without adding additional epochs overhead ($>$ 4X reduction in train steps in $\star$ABCI-2048, where the same techniques are applied) and, alternatively, - slightly less efficiently - to 4,096 with 1.4X extra epochs ($>$ 5X reduction in train steps in $\star$Fugaku-16384).}

Since the accuracy could not reach the target using a batch size larger than 4,096 even after \changes{extensive hyperparameter tuning}, model parallelism is necessary to scale beyond 4,096 processors. Therefore, a hybrid approach utilizing data- and model-parallelism is implemented for CosmoFlow on Fugaku. Model parallelism 
%in TensorFlow 
is implemented by extending Mesh TensorFlow so that multi processing can be applied for both data and model parallelism, and the hybrid parallelism is applied to Conv3d layers\changes{, that have a high spatial locality,} by partitioning input tensors spatially in two dimensions (Figure~\ref{figure:model-parallelism}).
\Lukas{What does "multi processes of both data and model parallelism are enabled, and applied to Conv3d layers" mean? Not clear to me}

\changes{To compare the hybrid data-and-model parallelism to pure data parallelism, Figure \ref{figure:fugaku-time-2epochs} shows a breakdown of the elapsed time for the first two epochs in CosmoFlow. The communication share increases as the number of CPUs increases in both cases. But whereas the scalability of data parallelism is limited mainly by the increase of the number of epochs (Figure \ref{figure:fugaku-epoch-time}/\ref{fig:analysis-cosmoflow-benchmark-overview}b), that of the hybrid parallelism is constrained by the increase of communication time (Figure \ref{sfigure:fugaku-time-2epochs-model-parallel}).}
As a result, the hybrid parallelism enabled scaling the number of CPUs in the submissions up to 8,192 with 4x4 spatial partitioning for closed division and 16,384 with 4x1 spatial partitioning for open division ($2.62 \times$ and $1.98 \times$ speedup of training throughput by model-parallelism compared to Fugaku-512, \mlhpc{resp.} an extrapolation thereof to batch size 4,096, \changes{Figure \ref{fig:analysis-cosmoflow-benchmark-overview}(c))}.
There is still room for improving the scaling efficiency of the hybrid parallelism further by reducing the communication overhead. 
\changes{In particular, we expect that parallelizing the halo-exchange of the Conv3D-layers within the model-parallel group with the evaluation of the layer in the non-halo region will give a significant speedup.}

\subsubsection{ABCI} 
\label{sssec:abci}

For both of the benchmarks, data reformatting by compressing and archiving multiple files was effective to reduce data staging time. Data shuffling was applied only among intra-node GPUs after each epoch, since the dataset is too large to fit on local storage and a partial dataset is shared only intra-node after data staging. \Lukas{check comment here}

For CosmoFlow, the following performance optimizations were applied to improve training and evaluation throughput: (1) improve data loader bandwidth using NVIDIA Data Loading Library (DALI), (2) apply mixed-precision training, (3) increase validation batch size. 
\changes{When applying the three optimizations sequentially on a single GPU, DALI is effective for both training (1.26X speedup) and evaluation (1.69X), adding mixed precision is effective only for training (1.77X), and increasing validation batch size further improves evaluation (3.03X).}
\mlhpc{With data-parallelism,} training time scales up to batch size 512 at a local batch size one for closed division.
For open division, after the hyperparameter tuning techniques mentioned in 
section~\ref{sssec:fugaku} were applied, batch size 2,048 with local batch size one was optimal.
Using a larger batch size than 2,048 did not achieve additional speedup because network bandwidth degraded due to congestion and the number of epochs increases significantly (Figure~\ref{figure:fugaku-epoch-time}). \changes{We have already seen a manifestation of this in the weak-scaling analysis of ABCI-512 and $\star$ABCI-2048 in subsection \ref{sec:compute-time-analysis} when comparing training and evaluation scaling efficiency (Figure \ref{fig:analysis-cosmoflow-benchmark-overview}c).}

For DeepCAM, page-locked memory (a.k.a pinned memory) is used to improve memory bandwidth, and four additional worker processes were forked for data loading to improve I/O bandwidth. Hyperparameters were tuned to reduce the number of epochs to convergence. %For DeepCAM, 
Training time scales up to batch size 2,048 with local batch size two for closed and open divisions after hyperparameter tuning. Especially tuning the warmup steps was effective to reduce the number of epochs to convergence. For the open division submission, the Gradient Skipping (GradSkip) technique~\cite{cac}
was applied that avoids updating weights in some layers in the training process, by finding layers which have little effect on accuracy, based on automatic analysis of the content of data during training. 
\changes{Effectively, GradSkip skips the gradient calculation of these layers in the backward pass. For our submissions, we apply it in two steps. After an average of 53.8\% train steps, we skip the first 62 out of 301 layers (12.4\% speedup) and after another 20.1\% train steps, we skip the first 260 layers (22.1\% speedup over no GradSkip).} 
\subsubsection{Cori/Cori-GPU}
\label{sssec:cori} 

Submissions on the Cori supercomputer at NERSC utilized both the primary KNL partition as well as the Cori-GPU testbed. System details are available at~\cite{cori,corigpu}.

CosmoFlow was trained on Cori KNL on 512 nodes in the closed division and 1024 nodes in the open division. For the open division submission, an additional learning rate decay \mlhpc{by a factor of 0.5 at 96 epochs} was added \mlhpc{to the schedule in section \ref{ssec:benchmark-suite-cosmoflow}} in order to enable convergence at global batch size 1024. The implementation used Intel-optimized TensorFlow with MKL-DNN for optimized performance on the Intel processors. Runtime settings for inter- and intra-parallelism threads, OpenMP threads, and affinity were tuned for maximal throughput. Shifter containers were used to launch training, which prevented scalability issues in shared library loading from the parallel file system at scale. These results show that large CPU systems like Cori can still be useful for training computationally-expensive deep learning models.

On an 8-node Cori-GPU system (64 V100 GPUs), Horovod with NCCL-based allreduce was used to achieve efficient data-parallel training. Additionally, node-local SSDs were used to store local partitions of the full dataset. The staging time from the Cori scratch filesystem to the node-local SSDs was considerably longer than other submissions with data-staging, indicating there is further room for optimization.
DeepCAM was similarly trained on the Cori-GPU system utilizing 64 V100 GPUs. The implementation in PyTorch utilized the NVIDIA Apex library for automatic mixed precision and used NCCL for optimized distributed data-parallel training.

\subsubsection{Piz Daint}
\label{sssec:piz-daint}
\begin{figure}[htbp]
\vspace{-1mm}
    \centering
    \includegraphics[width=0.65\linewidth]{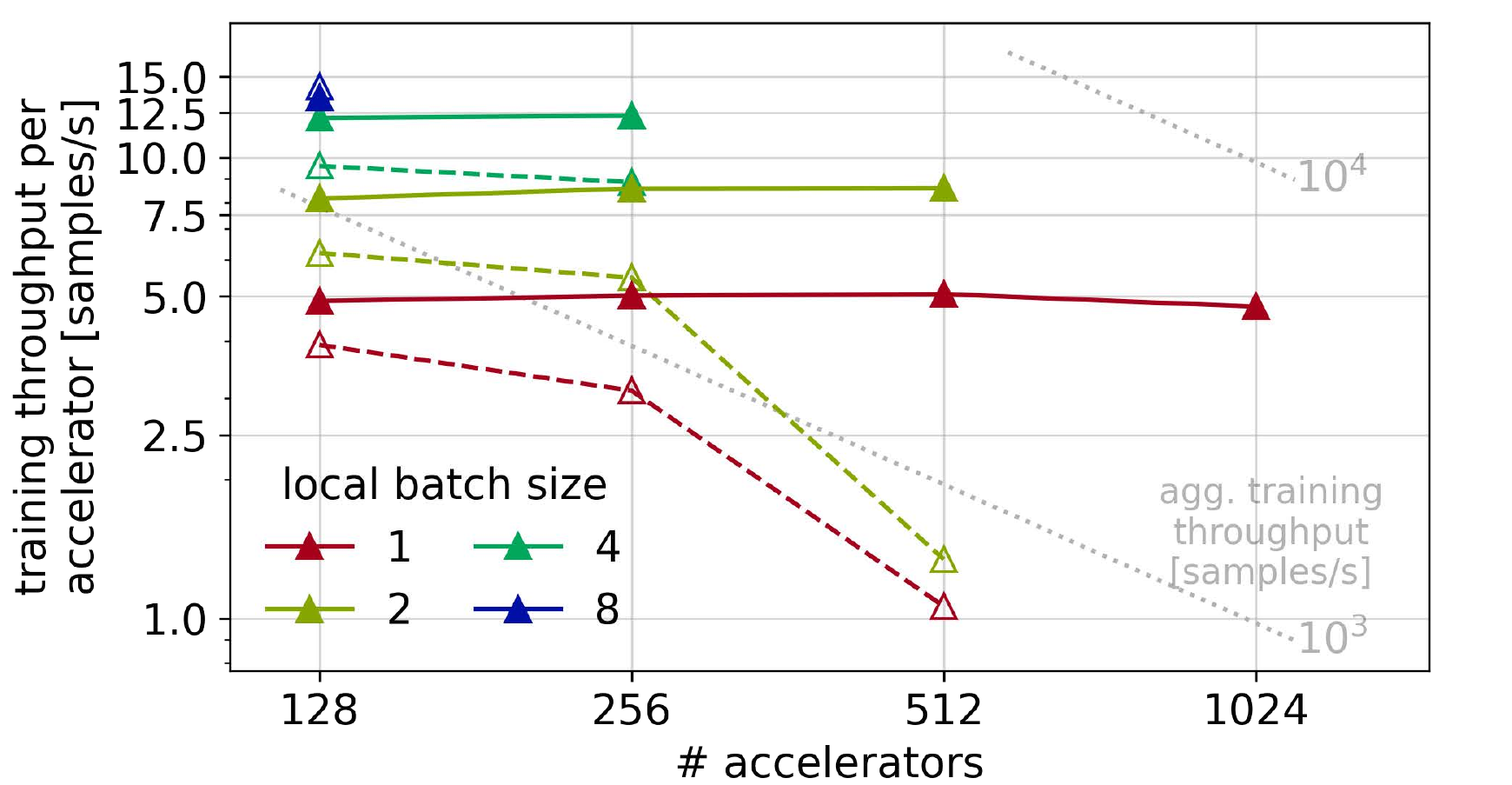}
    \vspace{-3mm}
    \caption{Weak-scaling of training throughput for CosmoFlow on Piz Daint before (hollow symbols, dashed lines) and after (filled symbols, solid lines) \changes{applying fine-grained communication} in the batch size region 128-1024.%Hovorod/NCCL-\changes{scheduling-}optimizations in the batch size region 128-1024.
    }
    \label{fig:piz-daint-training-throughput}
    \vspace{-2mm}
\end{figure}

% \Murali{reduce the text}
Submissions on Piz Daint~\cite{pizdaint} at CSCS focused on two data-parallel configurations in the closed division of CosmoFlow with 128 and 256 GPUs, one GPU per node. Sarus \cite{benedicic2019sarus}, a container engine with near-native performance for Docker-compatible containers, was used to rapidly test and tune distributed training with Horovod and NCCL for fine-grained communication to obtain near optimal weak scaling in the range of 100-1000 nodes (Figure~\ref{fig:piz-daint-training-throughput}). A low cycle time, tensor fusion threshold and the usage of the hierarchical, tree-based all-reduce implementation proved to be key to achieve this performance. \changes{Single node optimizations within Tensorflow had a comparatively smaller effect on training throughput, except for tuning intra-/inter-op parallelism threads \cite{wang2020exploiting}.}

To find the optimal batch size at a fixed node count, throughput scaling due to increased \changes{local batch size and, thus,} GPU data-parallelism has to be traded off against epoch scaling (Fig. \ref{fig:analysis-cosmoflow-benchmark-overview} b). Specifically, Figure \ref{fig:piz-daint-training-throughput} shows that the throughput ratio for a local batch size 4, 2, and 1 relative to a maximum local batch size of 8 which is 91\%, 64\%, and 35\% %that fits on the P100's memory 
and roughly coincides with the strong scaling efficiency \changes{at a batch size of 1,024}. 
Due to the \mlhpc{particular} closed division epoch scaling in Figure \ref{fig:analysis-cosmoflow-benchmark-overview}b, especially the strong slope starting from batch size 256 to 512 and a measured value of $59$ epochs for batch size 128, a local batch size of 2 turns out to give the fastest time to solution for both configurations.
To further reduce it, %we expect an
\changes{we expect an} alternative approach to data-parallelism, which only gives another 11\% improvement \changes{by doubling the resources}, to be more efficient.

Curiously, Figure \ref{fig:analysis-cosmoflow-benchmark-overview} (c) shows faster than ideal throughput scaling from 128 to 256 GPUs, +8\% compared to what is expected for training and +167\% for evaluation. %, respectively. 
\changes{So the time to solution is 12\% faster on 256 GPUs than expected based on the epoch scaling.} 
This is a result of caching the data set in RAM with 256 nodes, whereas at 128 nodes, parallel file system I/O (measured in section \ref{sec:workload-characterization}) is a bottleneck %, which directly impacts evaluation and 
which could be alleviated using near-compute storage. 

In summary, %we have identified 
fine-grained communication together with the addition of near-compute storage are identified as key optimizations for CosmoFlow on Piz Daint.

\section{Workload Characterization}
\label{sec:workload-characterization} 

 In this section, we present \mlhpc{techniques to characterize the benchmarks in terms of memory, network and I/O performance metrics}. 
\changes{This is motivated by the fact that these metrics parameterize extended roofline models \cite{cardwell2019extended} that allow to characterize future MLPerf HPC submissions with respect to system capabilities and
identify remaining optimization potential in submissions both at the hardware- and software-level. This will complement the information available from the high-level logs (section \ref{sec:results}). 
}
 It is to be noted that these metrics were captured \changes{separately from v0.7 submissions in additional runs, with 2 epochs per run} \mlhpc{in a purely data-parallel setup unless noted otherwise}.
For CosmoFlow, we used  \changes{a quarter of the full dataset (for each training and evaluation)} %(65,536 training, 16,384 validation samples)  
and a local batch size of 1, %\revision{How many CPUs/GPUs are the experiments run on?}
while for DeepCAM, we used all data samples %(121,266 for training, 15,158 for validation) 
and a local batch size of 2.

\begin{table*}
\small
\caption{\mlhpc{Workload characterization: memory bandwidth (single CPU/GPU), network and per-worker I/O bandwidth measurements}}
\label{tab:perf-metrics-all}
\footnotesize
\centering
% \resizebox{1.9\columnwidth}{!}
{
    \begin{threeparttable}
  \begin{tabular}{p{1.1cm}p{1.1cm}|lp{1.8cm}|llp{1.8cm}p{1cm}|lp{1.2cm}}
    \toprule
    \textbf{Benchmark} & \textbf{System}                  & \textbf{Memory Tool}    & \textbf{Memory BW (GB/sec)} & \textbf{Network Tool} & \textbf{\# units} & \textbf{Network BW (GB/sec)} & \textbf{Size (MB)} & \textbf{I/O Tool} & \textbf{I/O BW (GB/sec)} \\
    \midrule
    %  CosmoFlow         & ABCI \changes{(mixed precision)} & Nvprof           & 335.4                & Horovod TL    & 512 GPUs          & 3.41                 & 19.97              & Nvprof        & 1.65 \\
                    %   & \changes{ABCI (FP32)}            & \changes{Nvprof} & \changes{427.1}                                                                                                             \\                                                                 
          CosmoFlow         & ABCI\tnote{\textdagger}         & Nvprof           & 335.4                & Horovod TL    & 512 GPUs          & 3.41                 & 19.97              & Nvprof        & 1.65 \\
                       & Fugaku\tnote{\textdagger}       & Perf             & 110.8      & Mpitrace              & 512 CPUs          & 0.75                 & 21.71              & Timer-based   & 2.57 \\       
                       & Piz Daint                        & Nvprof          & -                    & Horovod TL    & 256 GPUs          & 1.86                 & 2.21               & Darshan       & 0.51 \\                     
                       & Summit                           & Nsight           & 233.1                & Horovod TL    & 510 GPUs          & 2.24                 & 22.0               & Darshan       & 1.46 \\
                       & ThetaGPU                         & Nsight           & 194.5                & Horovod TL    & 128 GPUs          & 1.95                 & 15.20              & Darshan       & 1.98 \\                
     \hline                                                                                                                                                                                          
     DeepCAM           & ABCI                             & Nvprof           & 153.1                & Timer-based   & 512 GPUs          & 3.73                 & 37.77              & Darshan       & 2.36 \\          
                       & Summit                           & Nsight           & 254.7                & Timer-based   & 510 GPUs          & 4.50                 & 225.0                                     \\          
    \bottomrule
  \end{tabular}
    \begin{tablenotes}
    \item[\textdagger] mixed-precision used on ABCI (memory bandwidth with FP32-training: 427.1 GB/sec), no model-parallelism used on Fugaku measurements
    \end{tablenotes}
    \end{threeparttable}
}
\vspace{-.25in}
\end{table*}

\vspace{-1mm}
\subsection{Memory Bandwidth}
We measure memory traffic of these benchmark implementations to estimate how much bandwidth is used for memory reads and writes to the off-chip DRAM on respective systems. More concisely, we measure the accelerator memory bandwidth 
aggregated across the system. \Lukas{? Not for single compute unit?} Global memory bandwidth is usually influenced by the underlying cache implementations and may not reflect the memory traffic in its entirety. Hence, we measure DRAM read and write throughput.
Table \ref{tab:perf-metrics-all} lists the average bi-directional bandwidth (read and write) 
across different systems. 

On ABCI, we used Nvidia \texttt{Nvprof} to calculate the average memory bandwidth of all kernels based on the elapsed time for each CUDA kernel and the memory bandwidth between L2 cache and HBM memory.
Since Fugaku does not have GPUs, we used \texttt{Perf}
\cite{perf} to extract read and write memory bandwidths measured at 1ms intervals 
and the average bandwidths are calculated for each. While \texttt{Nvprof} measures the bandwidth of CUDA kernel time only, \texttt{Perf} measures the bandwidth of the training interval at regular intervals. 
 On ThetaGPU and Summit we used Nvidia \texttt{Nsight compute} \cite{nsight} to extract the memory bandwidth of all kernels using the metric \texttt{dram\_\_bytes.sum.per\_second}. 

\textit{Observations:}
\changes{
Using an additional FP-32 memory bandwidth measurement on ABCI (Table \ref{tab:perf-metrics-all}$^{\dagger}$), we observe that the memory bandwidth of CosmoFlow on Fugaku is 3.85X smaller than on ABCI without mixed precision, even though the maximum effective memory bandwidth in a stream benchmark is similar for the two.} 
\mlhpc{Taking into account the throughput relations at 512 compute units in the submission results (Figure \ref{fig:analysis-cosmoflow-benchmark-overview}c) and the 1.4X speedup of mixed-precision training over FP-32 (section \ref{sssec:abci}), we can compare the systems in their caching efficiency. It turns out that FP32 on ABCI has a similar (90 \%) memory traffic per train step as Fugaku, whereas mixed-precision training transfers only 50 \% as much memory, which underlines its cache-friendly implementation. }

\begin{figure}[htbp]
    \subfigure[CosmoFlow : GPU]{\includegraphics[width=0.48\linewidth]{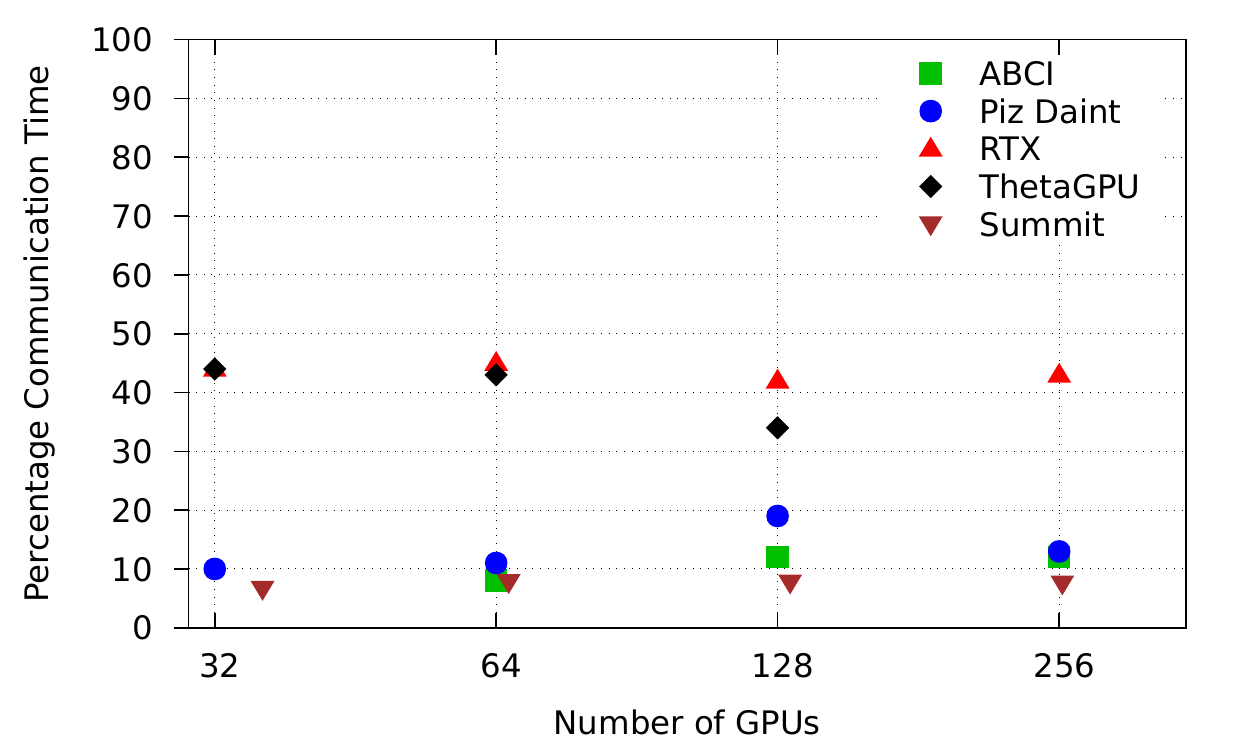}}
    \subfigure[DeepCAM : GPU]{\includegraphics[width=0.48\linewidth]{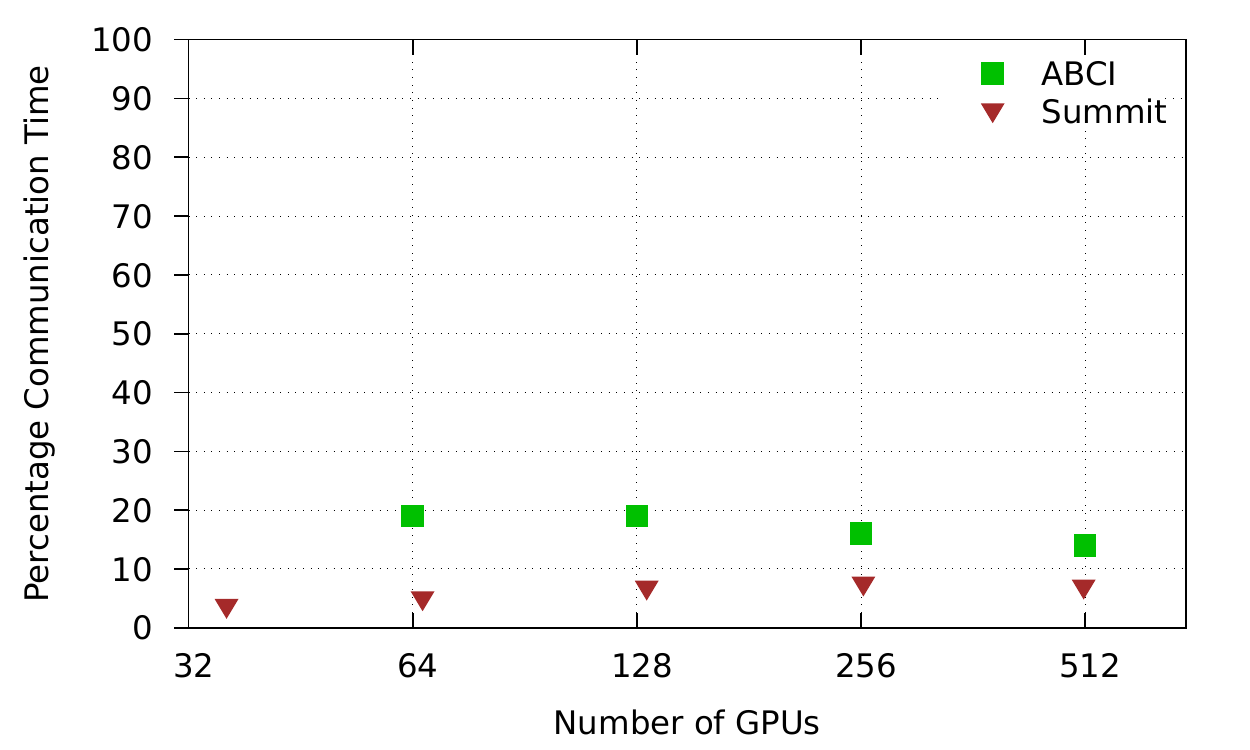}
    
    \label{sfigure:deepcam-abci}}
    \vspace{-2.5mm}
     \caption{Communication patterns in CosmoFlow and DeepCAM applications. 
     }
    \label{figure:collcomm}
    \vspace{-.2in}
\end{figure}

\subsection{Network Bandwidth}

To understand \mlhpc{the networking} %this
behavior, we profiled the heavy collective communication calls, \mlhpc{incl.} All\_Reduce operations, across \mlhpc{all} implementations. 
Figure~\ref{figure:fugaku-time-2epochs} and ~\ref{figure:collcomm}(a) presents the share of time spent in communication for CosmoFlow
and Figure~\ref{figure:collcomm}(b) shows similar measurements for DeepCAM. % on ABCI and Summit.

Since CosmoFlow's reference implementation uses Horovod, we used \texttt{Horovod Timeline} \cite{hvdtimeline} to obtain the average network bandwidth for collective communications
% on systems ABCI, Piz Daint, ThetaGPU and Summit 
as \texttt{mpitrace} \cite{mpitrace} was unable to correctly capture overlapping communication and computation.
The average network bandwidth is calculated based on the NCCL time obtained from the Horovod timeline, excluding the waiting time for data fusions. DeepCAM 
uses NCCL communication through \texttt{NVIDIA Apex} \cite{apex} in the reference implementation.
Since Horovod timeline and mpitrace cannot be used here, 
synchronization operations and timers are inserted before and after the collective communications of NVIDIA Apex to measure the communication time. Then we calculate the average communication bandwidth from the amount of actual data transferred.
On Fugaku, as Mesh-TensorFlow was used for CosmoFlow, 
Horovod timeline cannot be used here
and we use mpitrace and \texttt{mpiP} \cite{mpip} together to calculate the average communication bandwidth (results in Table \ref{tab:perf-metrics-all}).

\textit{Observations:}
Between 10-40\% of total application time is spent in collective communications as we scale across GPUs. % on the evaluated systems.
\mlhpc{On Fugaku, we observe that} for model-parallelism the scalability of the computation time is lower than that for the data-parallel execution. This is because model parallelism is applied only to the Conv3d layer (Figure \ref{figure:model-parallelism}).
Also, the communication time in absolute terms grows \mlhpc{faster} as the degree of model parallelism is increased \mlhpc{than with data-parallelism, where it is roughly constant}. This is due to the communication overhead caused by the data transfer in the halo region when performing spatial partitioning in the Conv3d layer.
\changes{Lastly, we observe that the small message size for Piz Daint %is significantly smaller compared to other systems, which 
is a direct consequence of the fine-grained communication optimization described in section \ref{sssec:piz-daint}.}

\subsection{I/O Bandwidth} 

As the MLPerf HPC benchmarks have massive input data sets, it is important to understand the I/O performance.
Table \ref{tab:perf-metrics-all} shows the average I/O bandwidth per worker on different systems. 
We used \texttt{Darshan} \cite{darshan} to get the average I/O bandwidth that captures all I/O-related activity, such as types and number of files and aggregate performance combined with shared and unique files worked by all ranks on certain systems. On ABCI, Darshan cannot measure the I/O volume accurately since the implementation of CosmoFlow used NVIDIA DALI which partly performs mmap-based I/O. Hence, we used \texttt{Nvprof} to measure the time of the kernel (\texttt{TFRecordReader}) that is performing I/O to calculate the average I/O bandwidth. 
On Fugaku, we insert timers before and after the data loads, and calculated from the elapsed time and the amount of data.

\textit{Observations:}
From Table~\ref{tab:perf-metrics-all}, it can be observed that the measured I/O bandwidth is similar \changes{for systems with on-node storage\footnote{This excludes Piz Daint (cf. section \ref{sssec:piz-daint}) as well as Frontera-RTX with 64 GPUs in the submission due to insufficient on-node SSD capacity.}} and we can expect that it is high enough to hide I/O behind computation. For example, for DeepCAM on ABCI, I/O bandwidth is 2.36 GB/s per process, using 256 processes with a full training dataset consisting of 7.7 TB. In this case, estimated I/O time per epoch is 7,700 GB / 256 processes / 2.36 GB/s = 12.8 seconds, while measured average run time per epoch that includes the I/O time is 99.6 seconds. 

\section{Key insights and conclusion}

\changes{To summarize, we presented MLPerf HPC, a benchmark suite aimed at representative scientific machine learning applications with two applications, CosmoFlow and DeepCAM.

We presented the results of initial submissions from leadership supercomputers and developed a framework for the systematic analysis of time to solution in terms of different components from data staging, algorithmic convergence and system \mlhpc{compute} throughput. This serves the critical need in the HPC community to understand large-scale scientific ML workloads from a holistic perspective. Furthermore, we introduced a set of techniques for workload characterization in terms of I/O, memory and network performance metrics to enable the parameterization of extended roofline models and, thereby, relate MLPerf HPC application performance to hardware capabilities in future rounds. 
}

\changes{The key insights from this round are:
\begin{itemize}[leftmargin=*,topsep=1pt]
    \item Data staging adds a highly variable overhead across applications %that is high variation 
    ($<1 \%$ to $20 \%$ of time-to-train) that depends % and depends 
    on both model and dataset characteristics. Where effective, compression and archiving of multiple files together should be done, %is extremely beneficial, 
    especially on smaller systems, where convergence in general is achieved in fewer epochs. 
    \item Accelerated storage solutions like on-node SSDs are critical %pieces of the memory hierarchy 
    for I/O-performance with large datasets. The results for CosmoFlow showed a narrow range of scale where GPU-based systems operated most efficiently by being able to fit the dataset in RAM and not yet experiencing prohibitive overhead from epoch scaling. If RAM capacity is insufficient, selectively caching the evaluation dataset in RAM should be attempted and performance can be further enhanced by improving data loader bandwidth and restricting data shuffling to intra-node (DeepCAM).
    \item Mixed-precision training and increasing the validation batch size significantly increase compute performance
    \item Efficient scheduling of communication is crucial for both data- and model-parallelism - while fine-grained communication is required in some data-parallel frameworks, model-parallel scalability heavily depends on overlapping halo exchanges with local layer-wise computations.
    \item Scaling to large batch sizes is challenging, requiring model-specific techniques such as special learning rate schedules, data augmentation and disabling dropout-layers that can exhibit a complicated interplay with convergence. This reinforces the need for efficient strong scaling methods, such as the hybrid model-and-data parallelism %shown 
    on Fugaku.
\end{itemize}
}

\changes{
\textit{Future Work:}
In future releases of the benchmark suite, we aim to
add new benchmarks for greater diversity and coverage of scientific ML workloads, including state-of-art models such as transformers and graph neural networks, as well as consider applications from emerging focus areas such as AI-driven simulations.
We also plan to expand the set of collected metrics to enhance performance models and, thus, increase utility and relevance to HPC users.}

\section*{Acknowledgment}
% \textit{add acknowledgements} \\
This research was funded in part by the Argonne Leadership Computing Facility, which is a DOE Office of Science User Facility supported under Contract DE-AC02-06CH11357.

\balance
\bibliographystyle{IEEEtran}

\clearpage
\appendix[Artifact Description/Artifact Evaluation]
\thispagestyle{empty}

\sloppy
\section*{Summary of the experiments reported}
We ran MLPerf\textsuperscript{\tiny{TM}} HPC benchmarks, CosmoFlow and DeepCAM on several supercomputers such as Cori, Fugaku and Piz Daint with Tensorflow, Horovod and PyTorch. The details are listed in sections \ref{sec:benchmarking-process} and \ref{sec:results} in the paper.

The baseline implementations are available at \textit{\url{https://github.com/mlcommons/hpc}}. The logging package can be installed as the instructions listed in this URL.

\begin{lstlisting}

# Install the package into your python environment.
git clone -b hpc-0.5.0 https://github.com/mlperf-hpc/logging.git mlperf-logging
pip install [--user] -e mlperf-logging

# Test compliance of a specific mlperf hpc log file
python -m mlperf_logging.compliance_checker --ruleset hpc_0.5.0 $logFile

\end{lstlisting}

\paragraph{CosmoFlow}

The dataset we use for this benchmark comes from simulations run by the ExaLearn group and hosted at NERSC at \textit{\url{ https://portal.nersc.gov/project/m3363/}}. The latest pre-processed dataset in TFRecord format is in the \textit{cosmoUniverse\_2019\_05\_4parE\_tf} folder, which contains training and validation subfolders. There are currently 262144 samples for training and 65536 samples for validation/testing. The combined size of the dataset is 5.1 TB.
Example submission scripts are in \textit{scripts} and YAML configuration files are in \textit{configs} directories.
To run a code at NERSC, use \textit{sbatch -N 64 scripts/train\_cori.sh} and modify the scripts accordingly 
to run on other systems.

\paragraph{DeepCAM}
The dataset for this benchmark comes from CAM5 simulations and is hosted at NERSC. The samples are stored in HDF5 files with input images of shape (768, 1152, 16) and pixel-level labels of shape (768, 1152). The globus endpoint for the dataset is at
\textit{\url{https://tinyurl.com/3hnd2z9c}}.
The splitting scripts to split the dataset to train, test, and validate are under \textit{src/utils}.
Example submission scripts are at \textit{src/deepCam/run\_scripts}. These can be modified to run on other systems. 

The v0.7 submissions are hosted at \textit{\url{https://github.com/mlcommons/hpc_results_v0.7}}. There is a separate directory for each submitting organization with the details of its submissions. 
It contains the following three sub-directories:
\begin{itemize}
    \item \textbf{benchmarks}: This directory contains the benchmark implementations and manual for system setup used in that organization's submissions. In particular, the detailed steps including the scripts to configure a system and run the benchmark are mentioned in a README file in  \textit{\textlangle{}submitter\textrangle{}/\textlangle{}benchmark-name\textrangle{}/implementations/\textlangle{}submission-entry\textrangle{}/} for each submission. The code used by a submission is referenced from there and available from \textit{\textlangle{}submitter\textrangle{}/\textlangle{}benchmark-name\textrangle{}/implementations/\textlangle{}implementation-entry\textrangle{}/} (multiple submissions can use the same code) and the system used is detailed at \textit{\textlangle{}submitter\textrangle{}/systems/\textlangle{}submission-entry.json\textrangle{}}.  
    For example, \textit{\url{https://github.com/mlcommons/hpc\_results_v0.7/tree/main/Fujitsu/benchmarks/cosmoflow/implementations/fugaku\_16384xA64FX\_tensorflow\_open}} has all the steps used for the open division submission to CosmoFlow on Fugaku with 16,384 CPUs. The implementation is available from \textit{\url{https://github.com/mlcommons/hpc\_results_v0.7/tree/main/Fujitsu/benchmarks/cosmoflow/implementations/implementation\_fugaku\_open}} and the system description from \textit{\url{https://github.com/mlcommons/hpc_results_v0.7/blob/main/Fujitsu/systems/fugaku_16384xA64FX_tensorflow_open.json}}.
    \item \textbf{results}: This directory lists all the submission results of an organization as log files, with one folder at \textit{\textlangle{}submitter\textrangle{}/results/\textlangle{}submission-entry\textrangle{}/\textlangle{}benchmark-name\textrangle{}/} for each submission. 
    For example, \textit{\url{https://github.com/mlcommons/hpc_results_v0.7/tree/main/CSCS/results/daint_gpu_n256_tf2.2.0/cosmoflow}} contains the log files for the closed division submission to CosmoFlow on Piz Daint system with 256 GPUs.
    \item \textbf{systems:} Here the details of the system hardware and software for each submission 
    is listed in a JSON format. In the paper, Table \ref{tab:systemdetails} lists the overall configuration of the systems while Table \ref{tab:closed-results} lists that used for the v0.7 submissions.
    For example, \textit{\url{https://github.com/mlcommons/hpc_results_v0.7/tree/main/LBNL/systems}} lists out Cori and Cori-GPU configurations used for different submissions.
\end{itemize}

Further details on the directory structure of the v0.7 submission results can be obtained from /\textit{\url{https://github.com/mlcommons/policies/blob/master/submission\_rules.adoc}}.

The analysis of the v0.7 submissions presented in section \ref{ssec:results-analysis} can be reproduced for both CosmoFlow and DeepCAM (for DeepCAM not shown in the paper) with the framework available at \textit{\url{https://github.com/lukasgd/mlperf\_hpc\_results\_v0.7\_visualization}} (results are in the branches \textit{hpc\_results\_v0.7\_cosmoflow} and \textit{hpc\_results\_v0.7\_deepcam}).

The full instructions to run the workload characterization experiments in Section \ref{sec:workload-characterization} (memory, network and I/O measurements) on the systems in Table \ref{tab:perf-metrics-all} are provided on \textit{\url{https://github.com/memani1/mlperfhpc-workload-characterization}}.

\paragraph{Author-Created or Modified Artifacts}

\noindent Persistent ID: \textit{\url{https://github.com/mlcommons/hpc}}\\
Artifact name: MLPerf HPC Benchmark Suite Reference Implementation

\noindent Persistent ID: \textit{\url{https://github.com/mlcommons/hpc\_results\_v0.7}}
Artifact name: MLPerf HPC Benchmark Suite Submissions v0.7

\noindent Persistent ID: \textit{\url{https://github.com/lukasgd/mlperf\_hpc\_results\_v0.7\_visualization}}\\\
Artifact name: MLPerf HPC Benchmark Suite Submissions v0.7 Results Visualization 

\noindent Persistent ID: \textit{\url{https://github.com/memani1/mlperfhpc-workload-characterization}}\\\
Artifact name: Workload characterization of MLPerf HPC benchmarks 

%\end{Verbatim}

% \end{document}

\end{document}